\documentclass[]{fairmeta}

\title{\vspace{-1pt}StereoVGGT: A Training-Free Visual Geometry Transformer for Stereo Vision}

\author[]{Ziyang Chen}

\author[]{Yansong Qu}

\author[]{You Shen}

\author[]{Xuan Cheng}

\author[*]{Liujuan Cao}

\affiliation[]{Key Laboratory of Multimedia Trusted Perception and Efficient Computing, \\Ministry of Education of China, Xiamen University}

\contribution[*]{corresponding author}

\usepackage{amsmath,amsfonts,bm}

\def\eqref#1{equation~\ref{#1}}

\def\1{\bm{1}}

\DeclareMathAlphabet{\mathsfit}{\encodingdefault}{\sfdefault}{m}{sl}
\SetMathAlphabet{\mathsfit}{bold}{\encodingdefault}{\sfdefault}{bx}{n}

\usepackage{color}
\usepackage{epsfig}
\usepackage{graphicx}

\usepackage{booktabs}  %
\usepackage{tabularx}               %
\usepackage{ltablex}
\usepackage{tabularray}
\newcolumntype{C}{>{\centering\arraybackslash}X}
\usepackage{multirow}               %
\usepackage{diagbox}                %
\usepackage{hhline}                 %
\usepackage{color}                  %

\usepackage{booktabs}
\usepackage{extarrows}
\usepackage{makecell}
\usepackage{wrapfig}
\usepackage{colortbl}
\usepackage[table]{xcolor}
\usepackage{longtable}
\usepackage{tabularray}
\usepackage{fancyvrb}

\usepackage{adjustbox}
\usepackage{array}
\usepackage{floatflt}

\usepackage{amsmath,amsfonts,amssymb}
\usepackage{bm}
\usepackage{nicefrac}
\usepackage{microtype}

\usepackage{changepage}
\usepackage{extramarks}
\usepackage{fancyhdr}
\usepackage{setspace}
\usepackage{soul}
\usepackage{xspace}
\usepackage{multicol}

\usepackage{url}

\usepackage{enumerate}
\usepackage{enumitem}  %

\usepackage{makecell}

\usepackage{pifont} %

\usepackage{algorithm,algpseudocode}

\usepackage[symbol]{footmisc}
\renewcommand{\thefootnote}{\fnsymbol{footnote}}

\usepackage[most]{tcolorbox}

\usepackage{caption}
\usepackage{scalefnt}
\usepackage{fontawesome5}

\usepackage{titletoc}
\definecolor{citecolor}{HTML}{0071BC}
\definecolor{linkcolor}{HTML}{ED1C24}
\usepackage{hyperref}
\usepackage{url}
\usepackage{multirow}
\usepackage{marginnote}
\usepackage{xcolor}
\usepackage{colortbl}
\usepackage{amssymb}
\usepackage{array}
\usepackage{xspace}
\usepackage{makecell}
\usepackage[table]{xcolor}
\usepackage{graphicx}
\usepackage{lipsum}
\usepackage{wrapfig}
\usepackage{nicematrix}
\usepackage{booktabs}
\usepackage{subcaption}
\usepackage{enumitem}
\setlist[itemize]{leftmargin=*}
\usepackage{siunitx}    %
\usepackage{makecell}   %
\usepackage{calc} 
\usepackage{cleveref}
\crefname{figure}{Fig.}{Figs.}
\crefname{table}{Tab.}{Tabs.}

\usepackage{titlecaps}
\usepackage{wrapfig}
\usepackage{amsmath}
\usepackage{caption}
\usepackage{wasysym}

\usepackage{tabularx}
\usepackage{etoolbox}

\renewcommand{\paragraph}[1]{%
	\noindent\textbf{#1.}\noindent\xspace%
}

\definecolor{qcolor}{HTML}{536872}

\newcolumntype{P}[1]{>{\centering\arraybackslash}p{#1}}

\newcommand{\tablestyle}[2]{%
	\fontfamily{ptm}\selectfont%
	\let\itold\it%
	\def\it{\itold \fontfamily{ptm}\selectfont}%
	\setlength{\tabcolsep}{#1}\renewcommand{\arraystretch}{#2}\centering\kindatiny%
	\let\citeold\cite%
	\renewcommand{\cite}[1]{\normalfont\fontfamily{ptm}\selectfont\tiny\citeold{##1}}%
}
\newcolumntype{x}[1]{>{\centering\arraybackslash}p{#1pt}}
\newcolumntype{y}[1]{>{\raggedright\arraybackslash}p{#1pt}}
\newcolumntype{z}[1]{>{\raggedleft\arraybackslash}p{#1pt}}
\newcolumntype{w}{>{\centering\arraybackslash}p{18pt}}
\newcolumntype{u}{>{\centering\arraybackslash}p{20pt}}
\newcolumntype{a}{>{\centering\arraybackslash}p{16pt}}
\newcolumntype{Y}{>{\centering\arraybackslash}X}

\newcolumntype{L}[1]{>{\raggedright\arraybackslash}p{#1}}
\newcolumntype{R}[1]{>{\raggedleft\arraybackslash}p{#1}}

\titlecontents{section}
[1.5em] %
{\addvspace{-0.5pt}} %
{\bfseries\contentslabel{2.3em}} %
{\hspace*{-2.3em}\bfseries} %
{\bfseries\titlerule*[.5pc]{.}\contentspage} %
\titlecontents{subsection}
[3.8em] %
{\addvspace{-2.2pt}} %
{\contentslabel{2.3em}}
{\hspace*{-2.3em}}
{\titlerule*[.5pc]{.}\contentspage}

\abstract{
	Driven by the advancement of 3D devices, stereo vision tasks including stereo matching and stereo conversion have emerged as a critical research frontier. Contemporary stereo vision backbones typically rely on either Monocular Depth Estimation models or general-purpose Pre-trained Vision Models. Crucially, these models are predominantly pretrained without explicit supervision of camera poses. Given that such geometric knowledge is indispensable for stereo vision, the absence of explicit spatial constraints constitutes a significant performance bottleneck for existing architectures. 
	Recognizing that the Visual Geometry Grounded Transformer (VGGT) operates as a foundation model pretrained on extensive 3D priors, including camera poses, we investigate its potential as a robust backbone for stereo vision tasks. Nevertheless, empirical results indicate that its direct application to stereo vision yields suboptimal performance. 
	We observe that VGGT suffers from a more significant degradation of geometric details during feature extraction. 
	Such characteristics conflict with the requirements of binocular stereo vision, thereby constraining its efficacy for relative tasks. 
	To bridge this gap, we propose StereoVGGT, a feature backbone specifically tailored for stereo vision. By leveraging the frozen VGGT and introducing a training-free feature adjustment pipeline, we mitigate geometric degradation and harness the latent camera calibration knowledge embedded within the model. 
	StereoVGGT-based stereo matching network achieved the $1^{st}$ rank among all published methods on the KITTI benchmark, validating that StereoVGGT serves as a highly effective backbone for stereo vision. 
\vspace{-10pt}
}
\metadata[Website]{\url{https://stereovggt.github.io}}

\definecolor{lightgray}{rgb}{0.95, 0.95, 0.95}

\definecolor{baselinecolor}{gray}{.9}

\begin{document}
\maketitle
\renewcommand{\thefootnote}{\arabic{footnote}}
\setcounter{footnote}{0}

\begin{center}
	\includegraphics[width=\linewidth]{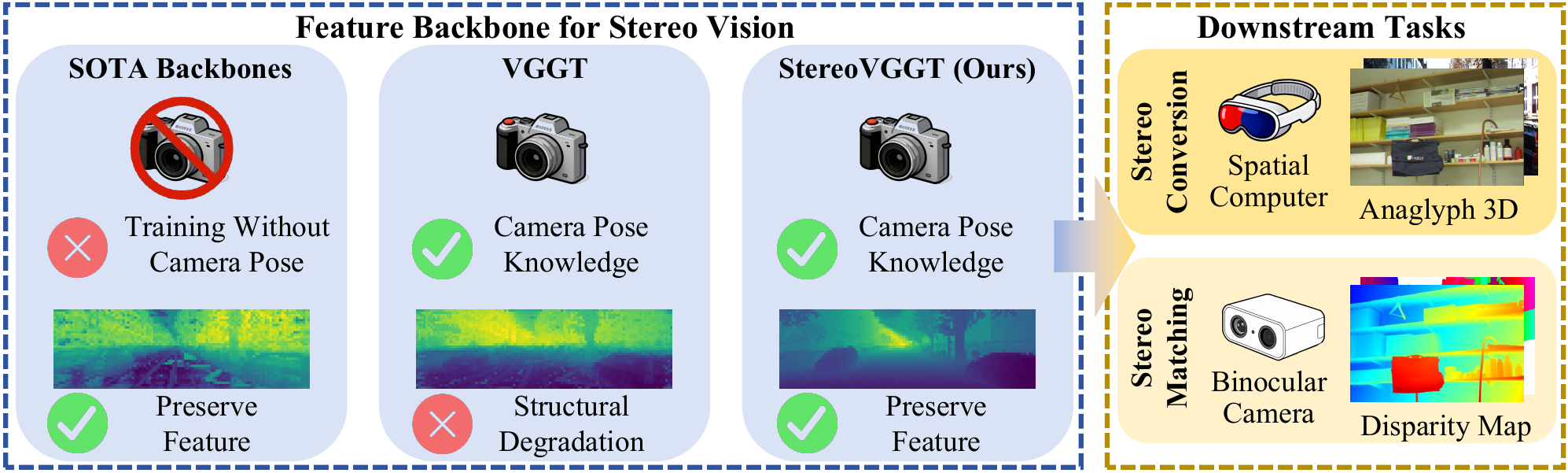}
	\captionsetup{type=figure}
	\captionof{figure}{Camera focal length serves as a determining factor in disparity estimation.Existing stereo vision backbones lack targeted learning of camera geometry and therefore largely neglect this critical prior.
	Given that VGGT is explicitly trained on 3D geometric priors such as camera parameters, we seek to exploit its inherent capacity for encoding camera representations. However, we observe that VGGT tends to excessively degrade the structural contours within the feature maps.
	This smoothing property is architecturally incompatible with the pixel-accurate alignment demands of stereo vision, creating a bottleneck for downstream stereo applications.
	StereoVGGT integrates the camera geometry knowledge encoded in VGGT while preserving robust feature representation capabilities. It can serve as a highly effective backbone for stereo vision.}
	\label{teaser}
\end{center}

\newpage
\section{Introduction}
\label{sec1}

Stereo vision constitutes a computational emulation of biological binocularity, serving as a fundamental pillar within the broader architecture of 3D vision perception \citep{sgm,eth3d,middlebury}. Stereo matching and stereo conversion serve as two core tasks within this domain. 
Stereo matching, also known as disparity estimation, aims to establish dense pixel-wise correspondences between rectified image pairs. Stereo conversion involves the synthesis of stereoscopic content from monocular images. Learning-based stereo conversion frameworks typically generate the corresponding right-eye view by leveraging estimated disparity as geometric guidance.

Disparity $d$ is inversely proportional to metric depth. Under the standard stereo imaging model, this relationship is expressed as $d = \frac{f \cdot B}{z}$, where $f$ denotes the focal length, $B$ denotes the baseline length, and $z$ denotes the metric depth. 
Inspired by this principle, current stereo matching and stereo conversion frameworks increasingly adopt Monocular Depth Estimation (MDE) models \citep{obukhov2025fourth,depthcrafter} and general-purpose Pre-trained Vision Models (PVMs) \citep{pvm-notion1} as their backbones for depth-aware feature extraction. 
However, from a geometric perspective, camera knowledge is also essential for stereo vision. 
Camera parameters explicitly encode the focal length $f$, which, according to the above relationship, directly determines the disparity $d$ for a fixed baseline and depth. 
Existing PVMs and MDE models are typically trained without explicit exposure to camera parameters or other 3D information. 
Therefore, the absence of camera knowledge limits the ability of existing stereo vision backbones to learn sufficient geometric priors, preventing them from fully exploiting scene geometry in downstream stereo vision tasks.

To demonstrate that current PVMs and MDE models lack sufficient camera knowledge, we conduct a comparative analysis by first extracting feature representations from a representative PVM (DINOv2), a representative MDE model (Depth Anything V2, DAv2), and the VGGT. 
Subsequently, following the methods in \citep{vggt,moge}, we implement a Levenberg-Marquardt solver to estimate the camera Field-of-View (FOV) \citep{fov} directly from these features. Detailed specifications of this solver are provided in the supplementary material.  
Finally, as illustrated by the bar charts in Fig. \ref{motivation}, the evaluation reveals that both DINOv2 and DAv2 exhibit substantial errors in camera FOV estimation. Conversely, the 3D foundation model VGGT, which is trained with explicit camera calibration supervision, delivers substantially better intrinsics estimation performance. These results suggest that both PVMs and MDE models fail to encode sufficient camera knowledge. In contrast, VGGT encodes camera knowledge substantially more effectively. This observation motivates us to leverage the frozen VGGT to enhance stereo vision performance.

\noindent
\begin{minipage}[t]{0.49\linewidth}
	\centering
	\includegraphics[width=\linewidth]{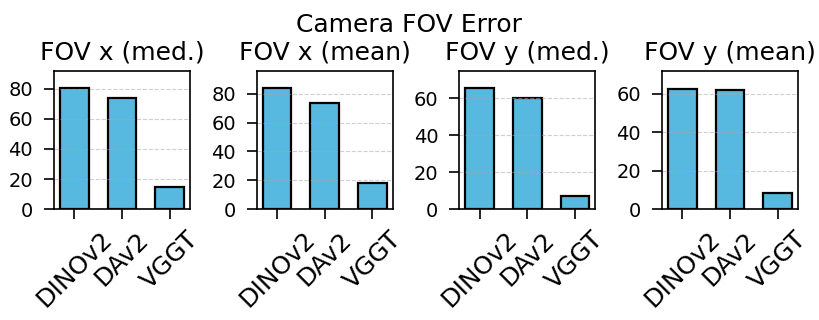}
	\captionsetup{type=figure}
	\vspace{-15pt}
	\caption{VGGT demonstrates a superior understanding of camera knowledge compared to SOTA models. The bar charts show the median and mean camera FOV errors across different methods. This analysis evaluates FOV estimated from extracted features on the ETH3D dataset.}
	\label{motivation}
\end{minipage}\hfill
\begin{minipage}[t]{0.49\linewidth}
	\centering
	\includegraphics[width=\linewidth]{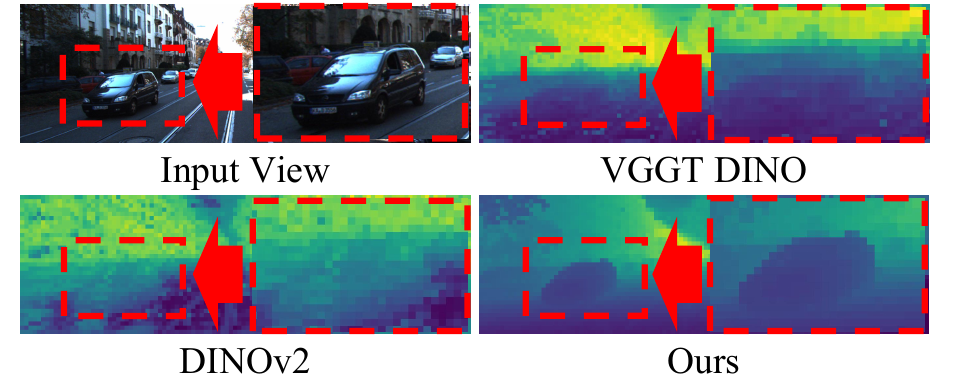}
	\captionsetup{type=figure}
	\vspace{-15pt}
	\caption{Feature maps visualization. The red bounding boxes highlight the vehicle contours extracted by each method. VGGT produces blurred vehicle contours, whereas DINOv2 and StereoVGGT retain them with complete fidelity.}
	\label{motivation2}
\end{minipage}
\vspace{5pt}

\begin{figure}[t]
	\centering
	\begin{minipage}[t]{0.49\linewidth}
		\centering
		\includegraphics[width=\linewidth]{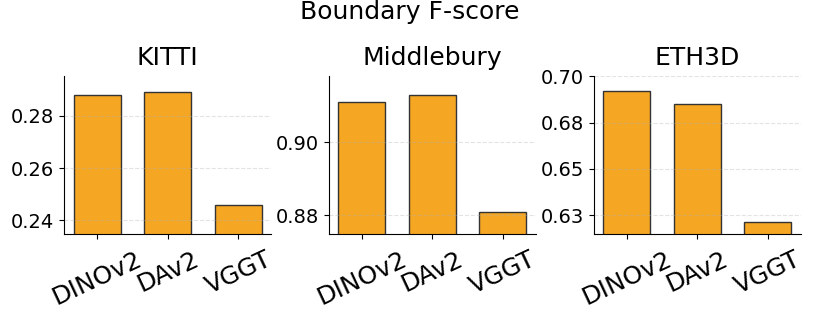}
	\end{minipage}\hfill
	\begin{minipage}[t]{0.49\linewidth}
		\centering
		\includegraphics[width=\linewidth]{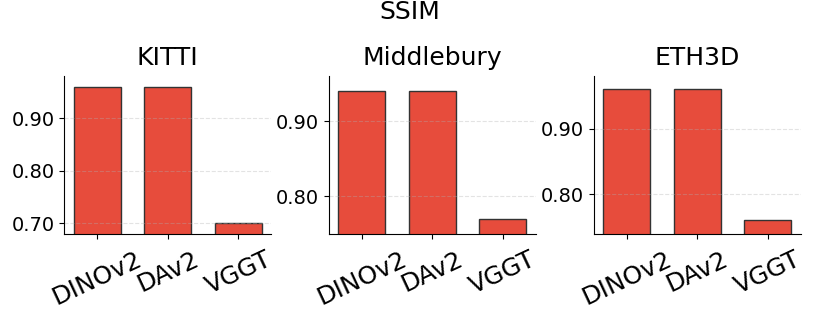}
	\end{minipage}
	\caption{VGGT suffers from structural degradation during its feature processing. The orange bar charts represent the Boundary F-score. Specifically, this metric quantifies the alignment between the boundaries of the feature map and those of the ground-truth disparity map.
		Concurrently, the red bar charts illustrate the SSIM values calculated between the extracted feature maps and the original input images across the ETH3D, KITTI, and Middlebury datasets. To ensure a fair comparison, all evaluated methods adhere to a consistent evaluation protocol; comprehensive details are provided in the supplementary material.}
	\label{motivation1}
\end{figure}

VGGT has been widely adopted as a feature backbone for various 3D vision tasks. However, empirical observations in the Experiments section demonstrate that its direct application fails to yield significant improvements in stereo vision performance. To understand why VGGT fails to provide immediate performance gains, we further investigate its intermediate feature representations. 
Specifically, we reshape the feature sequence extracted from the DINO encoder into spatial feature maps. 
As shown in Fig. \ref{motivation2}, features from original DINOv2 maintain clear geometric contours of the vehicle. In contrast, features from the DINO component of VGGT suffer from structural blurring, where the vehicle becomes nearly indistinguishable from the ground. 
For a more intuitive comparison, we evaluate the feature representations of DINOv2, DAv2, and VGGT using the Boundary F-score \citep{arbelaez2010contour} and Structural Similarity Index Measure (SSIM). The quantitative results are presented in Fig. \ref{motivation1}. Compared with DINOv2 and DAv2, VGGT exhibits consistently lower Boundary F-score and SSIM, indicating more severe structural degradation \citep{zeiler2014visualizing} within its feature representations. 

We attribute this phenomenon to the original design objective of VGGT, which is optimized for multi-view 3D reconstruction. To improve reconstruction robustness, the network tends to suppress local feature details, reducing error accumulation across views. However, unlike tasks that primarily benefit from high-level abstraction \citep{long2026}, stereo vision critically depends on preserving local geometric structures. Numerous stereo matching studies \citep{igev,mocha,selective} have demonstrated that maintaining or recovering fine-grained spatial details is essential for accurate disparity estimation. 

These observations suggest that the degradation of structural information limits the effectiveness of VGGT as a stereo vision backbone, despite its superior camera knowledge. To exploit the camera-specific knowledge encoded in VGGT while mitigating its structural degradation, we propose StereoVGGT, as summarized in Fig. \ref{teaser}. Specifically, StereoVGGT inherits the architecture and pre-trained weights of VGGT to preserve its camera-aware geometric priors, while introducing an entropy-based optimization strategy to merge the network parameters and restore structural information within the feature space. Our primary contributions are summarized as follows:
\begin{itemize}
	\item The absence of camera knowledge in existing backbones constitutes a bottleneck for improving accuracy in stereo tasks. Given that 3D foundation models are explicitly trained on such geometric data, we incorporate a frozen VGGT to supplement this knowledge.
	\item Building upon the entropy-minimization relationship observed between the weights of a 3D foundation model and an MDE model, we derive a data-free weight-merging strategy for our stereo vision backbone.
	\item We design a neck architecture that leverages VGGT's Frame Attention to modulate MDE features via feature-wise subtractive modulation, enabling a synergistic integration of the high-fidelity spatial details preserved by MDE models and the camera-specific geometric priors encoded in the 3D foundation model.
	\item The StereoVGGT-based stereo matching network achieves leading performance on the KITTI benchmark for non-occluded pixels, ranking $1^{st}$ at the time of submission. The StereoVGGT-based stereo conversion network achieves state-of-the-art performance on the Mono2Stereo and Inria 3D Movie datasets, demonstrating the potential of 3D foundation models for stereo vision.
\end{itemize}

\section{Related Work}\label{sec2}

\subsection{Relationship between Disparity and Depth}
In stereoscopic vision, the three-dimensional structure of the world is reconstructed from a pair of two-dimensional images. The fundamental cue for this process is binocular disparity---the horizontal positional difference between the projections of a scene point in the left and right images. 
Mathematically, for a 3D world point $P$, if its projections onto the left and right image planes have coordinates $x_l$ and $x_r$ respectively, the disparity $d$ is defined as $d=x_l - x_r$. 
This disparity $d$ is not a direct measure of distance but is geometrically inversely proportional to metric depth $Z$, the absolute distance from the observer, as described by the standard binocular imaging model: $Z = \frac{f \cdot B}{d}$, where $f$ is the focal length and $B$ is the baseline distance between the cameras. This equation provides a precise, quantitative mapping from image coordinates to a scaled metric interpretation of the scene, which is crucial for applications like robotic navigation and precise 3D reconstruction.

\subsection{Feature Backbone for Stereo Vision}
\subsubsection{Stereo Conversion} 
Mainstream methods typically follow a two-stage pipeline to generate stereoscopic views. In the first stage, a disparity map is estimated and used to geometrically warp the left view image. The second stage employs an inpainting model to generate the right view. The accuracy of stereo conversion heavily depends on precise disparity estimation \citep{stereodiffusion}. 

Recent leading approaches, such as StereoCrafter \citep{stereocrafter} and Mono2Stereo \citep{mono2stereo}, leverage advanced MDE models as backbones and integrate inpainting modules, achieving state-of-the-art (SOTA) performance. However, these MDE models are typically trained without explicit awareness of camera parameters. 
During the transformation from estimated depth to disparity, the absence of prior knowledge regarding focal length and baseline distance introduces geometric ambiguity. This lack of calibration awareness can lead to spatial information distortion.

\subsubsection{Stereo Matching}
Driven by large-scale datasets like ImageNet \citep{imagenet}, PVMs have demonstrated a strong capability for extracting stereo features. Employing these PVMs directly as feature backbones has shown superior performance compared to training feature extractors from scratch \citep{igev,mocha}. 
Recent studies \citep{los,monster} have shifted focus towards using MDE models as backbones for stereo matching. This trend is motivated by two factors: the improved accuracy of MDE models in estimating relative depth, and the inherent correlation between disparity and relative depth.

\subsubsection{Summary}
Current stereo vision backbones are typically trained without exposure to explicit camera calibration knowledge, which is fundamental to the depth-to-disparity conversion process. We contend that this omission of geometric calibration awareness constitutes a significant bottleneck, preventing these backbones from achieving more robust performance in diverse stereo vision scenarios.

\subsection{3D Foundation Models}
Recent advances in large-scale visual geometry learning have led to the emergence of 3D foundation models \citep{dust3r,omega}, which aim to learn a unified representation of scene geometry from diverse visual observations. Unlike conventional monocular depth estimation \citep{midas,unidepth} or multi-view stereo methods \citep{fdn,sat} that are designed for specific geometric tasks, these models jointly infer multiple scene attributes---including camera parameters, depth maps, point maps, and dense geometric correspondences---from arbitrary numbers of input images in a feed-forward manner, establishing a unified paradigm for visual geometry understanding.

Among them, VGGT \citep{vggt} represents a major milestone by introducing a transformer architecture that directly predicts camera intrinsics and extrinsics together with dense 3D representations from one or multiple images. 
Through alternating frame attention and global attention, VGGT learns geometry-aware features that implicitly encode camera calibration and global scene structure, achieving state-of-the-art (SOTA) performance across a wide range of geometric tasks while eliminating expensive geometric optimization during inference. 
Owing to the strong geometric priors, VGGT has rapidly evolved beyond general 3D reconstruction and have been successfully adopted as feature backbones for various downstream tasks, including feed-forward novel view synthesis \citep{idesplat}, and geometry-aware scene understanding \citep{mvggt}. Their common characteristic is the explicit learning of camera pose and global geometric consistency during large-scale pretraining, endowing the learned representations with substantially richer geometric semantics than conventional visual foundation models.

Despite the remarkable progress of VGGT, the applicability of 3D foundation models to stereo vision remains largely unexplored. Stereo vision fundamentally differs from holistic 3D reconstruction in that it requires preserving fine-grained local structures for accurate pixel-wise correspondence estimation \citep{croco,vitas}, whereas existing 3D foundation models are primarily optimized for scene-level geometric reasoning and camera estimation. 
Our preliminary investigations also reveal that the direct integration of VGGT into existing stereo vision frameworks does not yield the expected performance gains; in fact, such a straightforward substitution can be counterproductive, leading to performance degradation. This suggests that the priors learned by VGGT may conflict with the geometric correspondence requirements of stereo vision tasks in the absence of specialized adaptation. 

Consequently, although VGGT has demonstrated impressive transferability across numerous downstream tasks, there is still no evidence that 3D foundation models, including VGGT, can serve as effective feature backbones for stereo vision. 
This gap motivates us to systematically investigate whether the geometry-aware representations learned by modern 3D foundation models can be successfully transferred to stereo matching, and what architectural adaptations are required to fully exploit their potential.

\subsection{Structural Degradation}

In computer vision and signal processing, structural degradation refers to the progressive loss or distortion of the intrinsic geometric organization that defines an image or feature representation, including object boundaries, fine-scale textures, spatial continuity, and local structural relationships \citep{pei2019effects,wang2020deep,mocha}. Traditionally, this phenomenon is characterized in the image domain, where degradations such as blur, noise, and lossy compression corrupt the underlying spatial structure while preserving the overall scene semantics.

Recent advances in deep representation learning suggest that structural degradation also emerges within the latent feature space. As visual information propagates through successive network layers, feature representations are progressively transformed from low-level geometric descriptions into increasingly abstract semantic embeddings. Although this hierarchical abstraction substantially enhances semantic discrimination, it inevitably attenuates geometric details, leading to the erosion of structural cues encoded in the feature manifold. Consequently, deep feature maps gradually lose precise boundary localization, high-frequency textures, and fine-grained spatial correspondences relative to the original image, despite retaining rich semantic information \citep{zeiler2014visualizing,bau2017network}. In this work, we refer to this progressive attenuation of geometric fidelity within deep feature representations as structural degradation.

For many vision tasks \citep{long2025progressive,rethinking}, such structural degradation is not only acceptable but often advantageous, as suppressing low-level variations enables networks to learn more invariant and semantically meaningful representations. However, accurate disparity estimation fundamentally relies on preserving fine-scale structural information to establish reliable pixel-wise correspondences, as consistently demonstrated by numerous stereo vision studies \citep{igev,mochav2,los}. Excessive structural degradation compromises feature discriminability, thereby reducing matching precision. 

\subsection{Definition of Training-free}
The term training-free in this work specifically refers to the construction of the StereoVGGT backbone without performing gradient-based updates to the pre-trained model parameters using downstream training data \citep{taskvector,xu2024training}. In particular, the shared DINO weights inherited from VGGT and MoGe-2 remain frozen, and no labeled or unlabeled samples are used to optimize these backbone parameters. Instead, we perform a data-free coefficient selection procedure in which only the layer-wise mixing coefficients are iteratively adjusted to merge the two source models. The optimization signal is not the predictive entropy of model outputs, but the entropy of the block-norm distribution induced by the merged weights. By minimizing this entropy, the merging process encourages a more selective and less interfered parameter fusion while preserving the frozen nature of the original backbone parameters. Therefore, in our setting, training-free denotes data-free backbone construction without parameter learning on downstream data, rather than the absence of every optimization step in the overall pipeline.

\subsection{Levenberg-Marquardt Solver}
\label{solver}

To facilitate pose estimation and validate the implicit representation of camera parameters within our features, we use a specialized solver. 
Following the camera head of VGGT \citep{vggt}, our pose solver involves using a frozen DPT head \citep{dpt} to compute a 3D point cloud. The 3D points $(X, Y, Z)$ are then projected onto the 2D camera plane as $(u,v)$:
\begin{equation}
	\begin{bmatrix}u \\ v\end{bmatrix} = f \cdot \frac{1}{Z+s} \cdot \begin{bmatrix}X \\ Y\end{bmatrix}
\end{equation}
$f$ is focal length, $s$ is depth shift. 
We aim to minimize the objective function presented in, 
\begin{equation}
	\min_{f, s} \sum_{i=1}^{N} \left\| \frac{\mathbf{X}\mathbf{Y}_i}{Z_i +s} - \mathbf{u}\mathbf{v}_i \right\|^2
	\label{mini}
\end{equation}
\noindent
$\mathbf{X}\mathbf{Y}_i$ is the 2D coordinates of the $i$-th point, $\mathbf{u}\mathbf{v}_i$ is the observed 2D pixel coordinates of the $i$-th point. $Z_i$ is the depth value of the $i$-th point, $N$ means the total number of points. 
Followed by Moge \citep{moge}, the Levenberg-Marquardt Solver \citep{levenberg} is employed to iteratively estimate the focal length $f$ and depth shift $s$, 
\begin{equation}
	\begin{cases}
		f ^{\star} &= \frac{\sum_{i=1}^{N} \frac{\mathbf{X}\mathbf{Y}_i}{Z_i+s ^{\star} \cdot \mathbf{u}\mathbf{v}_i}}{\sum_{i=1}^{N} || \frac{\mathbf{X}_i\mathbf{Y}_i}{Z_i +s ^{\star}}|| ^2} \\
		s ^{\star}&= LM(arg\underset{s}{\operatorname{min}} || r(s) ||^2 )
	\end{cases}
	\label{pose-solver}
\end{equation}
To solve for $f$ and $s$ simultaneously, we employ an alternating optimization strategy. By first fixing $s$, the subproblem of solving for $f$ becomes convex, yielding the following optimal solution:
\begin{equation}
	f^\star(s) = \frac{\sum_{i=1}^{N} \mathbf{X}\mathbf{Y}_i^{\prime} \cdot uv_i}{\sum_{i=1}^{N} \left\| \mathbf{X}_i\mathbf{Y}_i^{\prime} \right\|^2}
\end{equation}
Here, $\mathbf{X}\mathbf{Y}_i^{\prime}=\frac{\mathbf{X}\mathbf{Y}_i}{\mathbf{Z}_i}$. The subsequent step involves back-substituting the expression for $f^{\star}(s)$ into the objective function to obtain a residual function that is a function of $s$ alone, 
\begin{equation}
	\mathbf{r}(s) = \begin{bmatrix}
		f^{\star}(s) \cdot \mathbf{X}\mathbf{Y}_1' - \mathbf{uv}_1 \\
		f^{\star}(s) \cdot \mathbf{X}\mathbf{Y}_2' - \mathbf{uv}_2 \\
		\vdots \\
		f^{\star}(s) \cdot \mathbf{X}\mathbf{Y}_N' - \mathbf{uv}_N
	\end{bmatrix}
	\label{residual}
\end{equation}
The final $s$ is estimated via the Levenberg-Marquardt algorithm, as formulated in Eq. (\ref{pose-solver}), from which the camera focal length $f=\begin{bmatrix}
	f_x \\ f_y
\end{bmatrix}$ is subsequently derived. $LM$ denotes the Levenberg-Marquardt solver. 
The camera Field-of-View (FOV) sought in this work can be analytically derived according to, 
\begin{equation}
	\begin{cases}
		FOV_x &=  2 \cdot arctan(\frac{W}{2\cdot f_x}) \\
		FOV_y &=  2 \cdot arctan(\frac{H}{2\cdot f_y})
	\end{cases}
	\label{fov-last}
\end{equation}
Here, $W$ and $H$ denote the width and height of the input image, respectively.

\section{Insights}
\subsection{Theoretical Benefits of VGGT's Understanding of Camera Knowledge for Stereo Vision}
\label{fov-insight}

As analyzed in Sec. \ref{sec1}, we argue that camera knowledge is important for stereo vision tasks. Such knowledge is potentially valuable for stereo vision, where disparity is not only a function of scene structure but is also intrinsically governed by camera calibration. In particular, under the standard stereo imaging model, the mapping between depth and disparity is directly modulated by the focal length and baseline. This suggests that a feature backbone with a stronger internal representation of camera parameters may provide more suitable geometric cues for downstream stereo reasoning. 
However, existing stereo vision backbones largely overlook this issue. Given that VGGT is trained on 3D data, including camera poses, we hypothesize that it may possess stronger camera knowledge than SOTA stereo vision backbones. 
To verify this hypothesis, we compute the camera FOV error according to the procedure described below.

\subsubsection{Camera FOV Measurement: Robust Camera Knowledge Representation in VGGT}

Camera field of view (FOV) \citep{lee2021ctrlc,patel2025camerahmr} provides a concise and physically meaningful proxy for evaluating whether a representation preserves intrinsic camera knowledge. Under the standard pinhole camera model, the horizontal and vertical fields of view are determined by the image resolution and focal length. Let $W$ and $H$ denote the image width and height, and let $f_x$ and $f_y$ denote the focal lengths along the horizontal and vertical directions, respectively. The ground-truth camera FOV is defined as
\begin{equation}
	\begin{cases}
		FOV_x &=  2 \cdot arctan(\frac{W}{2\cdot f_x}) \\
		FOV_y &=  2 \cdot arctan(\frac{H}{2\cdot f_y})
	\end{cases}
	\label{eq:fov_definition}
\end{equation}
This formulation makes FOV a direct function of camera intrinsics: a smaller focal length yields a wider viewing angle, whereas a larger focal length results in a narrower but more magnified observation.

To estimate camera FOV from a model representation, we follow the geometric procedure in Sec. \ref{solver}. We solve this problem using a Levenberg--Marquardt solver \citep{levenberg}, which yields the estimated focal lengths $\hat f_x$ and $\hat f_y$. The predicted FOV is then obtained analytically as
\begin{equation}
	\widehat{\mathrm{FOV}}_x = 2 \arctan\!\left(\frac{W}{2\hat f_x}\right), \qquad
	\widehat{\mathrm{FOV}}_y = 2 \arctan\!\left(\frac{H}{2\hat f_y}\right).
	\label{eq:fov_prediction}
\end{equation}

We define the camera FOV error as the absolute angular deviation between the predicted FOV and the ground-truth FOV derived from annotated camera intrinsics:
\begin{equation}
	e_x = \left| \widehat{\mathrm{FOV}}_x - \mathrm{FOV}_x \right|, \qquad
	e_y = \left| \widehat{\mathrm{FOV}}_y - \mathrm{FOV}_y \right|.
	\label{eq:fov_error}
\end{equation}
For the camera FOV evaluation, we report both the median and mean of $e_x$ and $e_y$ across all images. Lower values indicate that the underlying representation preserves camera knowledge more faithfully. 
We emphasize that this measurement primarily evaluates intrinsic rather than extrinsic camera knowledge. In particular, FOV estimation does not directly capture information about camera pose or inter-camera transformations. Nevertheless, accurate FOV recovery still provides strong evidence that the representation is sensitive to essential calibration cues and is therefore not purely appearance-driven. From the perspective of stereo vision, such sensitivity is especially important because focal-length-related geometric priors directly influence the depth-to-disparity relationship. Therefore, although FOV estimation does not fully characterize all aspects of camera geometry, it offers a principled and practically measurable criterion for evaluating whether a model has learned robust camera-aware representations.

Under the camera FOV evaluation described in Fig. \ref{motivation}, VGGT consistently exhibits substantially lower FOV estimation error than conventional pre-trained vision models and representative monocular depth estimation backbones, indicating that its internal features preserve richer camera-calibration information. 
This empirical finding complements its training background as a geometry-grounded foundation model and supports our decision to incorporate VGGT despite its feature-level structural degradation. 

\subsection{The Gap Between VGGT's Feature Degradation and the Requirements of Stereo Vision}

Taking VGGT as a representative example of a 3D foudation model, we observe that its intermediate feature representations exhibit severe structural degradation, as evidenced by the analyses in Figs. 3 and 4. Our analysis primarily focuses on two complementary metrics, namely the Structural Similarity Index Measure (SSIM) \citep{wang2004image} and the Boundary F-score \citep{arbelaez2010contour}. In the following, we describe how these metrics are computed on intermediate feature representations and discuss the observations derived from their analyses. 

\subsubsection{SSIM Measurement: Structural Similarity to the Input Image}
\label{ssim}

Since structural degradation is expected to reduce the resemblance between intermediate feature representations and the original image structure, we employ SSIM to quantify the structural similarity between the PCA-projected feature map and the input image.

The Structural Similarity Index Measure (SSIM) was originally proposed as a perceptual metric for evaluating image quality \citep{wang2004image}. Unlike traditional pixel-wise metrics such as Mean Squared Error (MSE) or Peak Signal-to-Noise Ratio (PSNR), SSIM assesses image similarity from the perspective of structural information preservation, under the assumption that the human visual system is highly adapted to extracting structural cues from natural images.

Given two image patches $I_1$ and $I_2$, SSIM evaluates their similarity through three complementary components: luminance consistency, contrast consistency, and structural consistency. Formally, SSIM is defined as,

\begin{equation}
	\text{SSIM}(I_1,{I_2})=
	\frac{(2\mu_{I_1}\mu_{I_2}+C_1)(2\sigma_{{I_1}{I_2}}+C_2)}
	{(\mu_{I_1}^2+\mu_{I_2}^2+C_1)(\sigma_{I_1}^2+\sigma_{I_2}^2+C_2)}
\end{equation}
where $\mu_{I_1}$ and $\mu_{I_2}$ denote the mean intensities of the two patches, $\sigma_{I_1}$ and $\sigma_{I_2}$ represent their standard deviations, and $\sigma_{{I_1}{I_2}}$ is the covariance between them. Constants $C_1$ and $C_2$ are introduced to stabilize the computation when the denominators approach zero \citep{wang2004image}. The resulting score ranges from $[-1,1]$, with larger values indicating stronger structural similarity.

Although SSIM was originally designed to compare signals residing in the same image space, recent studies have shown that it can also provide useful insights into the degree to which intermediate neural representations preserve the structural organization of the input image \citep{zhang2018unreasonable,wang2020deep}. Motivated by this observation, we employ SSIM as a quantitative indicator of structural preservation in the latent representations produced by different models. Specifically, we analyze the shared DINO-based visual encoder adopted by VGGT \citep{vggt}, MoGe-2 \citep{moge}, and DINOv2 \citep{dinov2}. Given an input image $I$, the encoder produces a sequence of visual tokens $F \in \mathbb{R}^{N\times1024}$, where $N$ corresponds to the number of tokens and $1024$ denotes the feature dimension. To recover the spatial arrangement of the latent representation, the token sequence is reshaped into a two-dimensional feature map $\hat{F}\in\mathbb{R}^{H\times W\times1024}$. Since SSIM requires image-like inputs, directly comparing the full 1024-dimensional feature tensor with the RGB image is not meaningful. Following the common visualization protocol adopted in representation analysis literature \citep{dinov2,dlnr}, we project the feature map onto a three-dimensional subspace using Principal Component Analysis (PCA). Let
\begin{equation}
	\Phi = \text{PCA}_3(\hat{F}),
\end{equation}
where $\Phi \in \mathbb{R}^{H\times W\times3}$ denotes the first three principal components, which capture the dominant spatial variance of the representation. The resulting PCA visualization is then normalized to the RGB range and treated as an image-like representation of the latent feature structure. Finally, the normalized PCA projection $\Phi$ is compared against the original input image $I$ using SSIM, we compute SSIM between $\Phi$ and $I$. 
As illustrated in Fig. 4, a higher score indicates that the PCA-projected latent representation retains stronger structural consistency with the input image, whereas a lower score suggests more severe structural degradation in the feature space, implying that the spatial organization of the scene is less faithfully preserved during feature extraction. We emphasize that this metric does not measure semantic similarity; instead, it quantifies the extent to which geometric boundaries, object layouts, and local spatial structures remain preserved within the intermediate feature space. Consequently, it serves as a practical indicator for assessing the degree of structural degradation exhibited by different models.

For a more intuitive comparison, we visualize the feature representations by projecting them into a single-channel feature map. As shown in Fig. 3, features from original DINOv2 maintain clear structural contours of the vehicle. In contrast, features from the DINO component of VGGT suffer from structural blurring, where the vehicle becomes nearly indistinguishable from the ground. 
Overall, we argue that the reduction in SSIM observed with VGGT reflects a severe structural degradation within the feature space. This phenomenon ultimately hampers the utility of VGGT as an effective backbone for stereo vision tasks.

\subsubsection{Boundary F-score Measurement: Structural Boundary Preservation}
\label{fscore}

While SSIM evaluates the preservation of global structural organization, the Boundary F-score focuses on whether geometrically meaningful boundaries remain distinguishable in the feature representation. Moreover, it directly performed in the original feature space, avoiding the ambiguity caused by dimensionality reduction techniques such as PCA.

Following classical boundary evaluation protocols in image segmentation and edge detection benchmarks \citep{arbelaez2010contour}, we measure the consistency between feature-derived boundaries and ground-truth disparity discontinuities. The specific computation procedure is as follows:

Step 1: Feature Boundary Extraction. 

Given a dense feature representation:
\begin{equation}
	F \in \mathbb{R}^{C \times H \times W},
\end{equation}
where $C=1024$, we first perform per-location feature normalization:
\begin{equation}
	\hat{F}(x)=
	\frac{F(x)}
	{\|F(x)\|_2+\epsilon},
	\label{eq:feature_norm}
\end{equation}
where $\epsilon=10^{-6}$ is introduced for numerical stability. 
This normalization removes feature magnitude bias and ensures that the following analysis focuses on spatial structure rather than activation scale.

Inspired by previous works that exploit spatial feature variations for boundary detection~\citep{xie2015holistically,liu2017richer}, we define feature boundary strength as the local feature variation:
\begin{equation}
	S_F(x)=
	\frac{1}{|\mathcal{N}|}
	\sum_{x'\in\mathcal{N}(x)}
	\left\|
	\hat{F}(x)-\hat{F}(x')
	\right\|_2 ,
	\label{eq:feature_boundary}
\end{equation}
where $\mathcal{N}(x)$ denotes the four-neighbor pixels around pixel $x$. 
Large values of $S_F(x)$ indicate strong spatial feature discontinuities, corresponding to potential object boundaries. 
The predicted feature boundary map is obtained as:
\begin{equation}
	B_F(x)=
	\mathbb{I}(S_F(x)>\tau_F),
	\label{eq:feature_binary_boundary}
\end{equation}
\begin{equation}
	\tau_F=P_{75}(S_F)
\end{equation}

Step 2: Disparity Boundary Extraction. 

Given the ground-truth disparity map:
\begin{equation}
	D \in \mathbb{R}^{H \times W},
\end{equation}
We extract geometric boundaries according to disparity discontinuities. 
The disparity gradient magnitude is computed as:
\begin{equation}
	S_D(x)=
	\|\nabla D(x)\|_2 .
	\label{eq:disparity_gradient}
\end{equation}
The ground-truth disparity boundary map is defined as:
\begin{equation}
	B_D(x)=
	\mathbb{I}(S_D(x)>\tau_D)
	\label{eq:disparity_boundary}
\end{equation}
\begin{equation}
	\tau_D=P_{75}(S_D)
\end{equation}

Step 3: Boundary F-measure Evaluation. 

We evaluate the agreement between feature boundaries and disparity discontinuities using F-score.

For a predicted boundary pixel $p$, it is considered correctly detected if a ground-truth boundary exists within a distance tolerance $r=3$:
\begin{equation}
	\min_{q\in B_D}
	\|p-q\|_2 < r .
	\label{eq:boundary_matching}
\end{equation}
The boundary precision and recall are computed as:
\begin{equation}
	P_b=
	\frac{TP}{TP+FP},
\end{equation}
\begin{equation}
	R_b=
	\frac{TP}{TP+FN}.
\end{equation}
The final boundary F-score is measured by:
\begin{equation}
	F_b=
	\frac{2P_bR_b}
	{P_b+R_b}.
	\label{eq:boundary_fscore}
\end{equation}
A higher $F_b$ indicates that the feature representation better preserves geometric boundaries.

To ensure a fair compairison, all disparity maps are downsampled to the resolution of the feature map for evaluation. It can be inferred from the above formulation that features with better-preserved spatial structure yield a higher $F_b$. A lower $F_b$ indicates more severe degradation of the spatial structure in the feature map.

\subsubsection{Summary: Structural Degradation in VGGT}

SSIM quantifies the preservation of local structural patterns, whereas the Boundary F-score evaluates the retention of geometric boundaries. Together, these metrics provide empirical evidence that enables us to quantitatively assess structural degradation in VGGT feature representations, as shown in Figs. 3 and 4. We attribute this phenomenon to the original design objective of VGGT, which targets multi-view 3D reconstruction. To improve robustness across numerous input views, the network tends to suppress local structural details, thereby mitigating error accumulation during reconstruction. However, this behavior is unfavorable for stereo vision, where preserving fine-scale geometric structures is essential for establishing reliable pixel-wise correspondences.

Our results suggest that alleviating this structural degradation substantially unlocks the potential of VGGT for stereo vision. As illustrated in Tab. \ref{our-vis}, StereoVGGT consistently achieves higher SSIM and Boundary F-score values than the original VGGT, indicating superior preservation of structural information and geometric boundaries. This improvement translates directly into stronger stereo matching performance. As shown in Fig. \ref{snap}, StereoVGGT ranked $1^{st}$ on the KITTI benchmark \citep{kitti} for non-occluded pixels at the time of submission.

\begin{table}[h]
	\centering
	\setlength{\tabcolsep}{0.01mm}
	\begin{tabular}{c|cc|cc|cc}
		\toprule[1.5 pt]
		\multirow{2}{*}{Method} & \multicolumn{2}{c|}{KITTI} & \multicolumn{2}{c|}{Middlebury} & \multicolumn{2}{c}{ETH3D} \\ \cmidrule(lr){2-3} \cmidrule(lr){4-5} \cmidrule(lr){6-7}
		& SSIM$\uparrow$       & F-score$\uparrow$      & SSIM$\uparrow$       & F-score$\uparrow$       & SSIM$\uparrow$       & F-score$\uparrow$      \\ \midrule
		VGGT                    & 0.68       & 0.25          & 0.75          & 0.88            & 0.74       & 0.62         \\
		Ours        & 0.97       & 0.29          & 0.94          & 0.93            & 0.96       & 0.69        \\
		\bottomrule[1.5 pt] 
	\end{tabular}
	\caption{Quantitative evaluation of structural degradation. F-score means Boundary F-score here, $\downarrow$ indicates that lower values are better.} 
	\label{our-vis}
\end{table}

\begin{figure}[t]
	\centering
	\includegraphics[width=\linewidth]{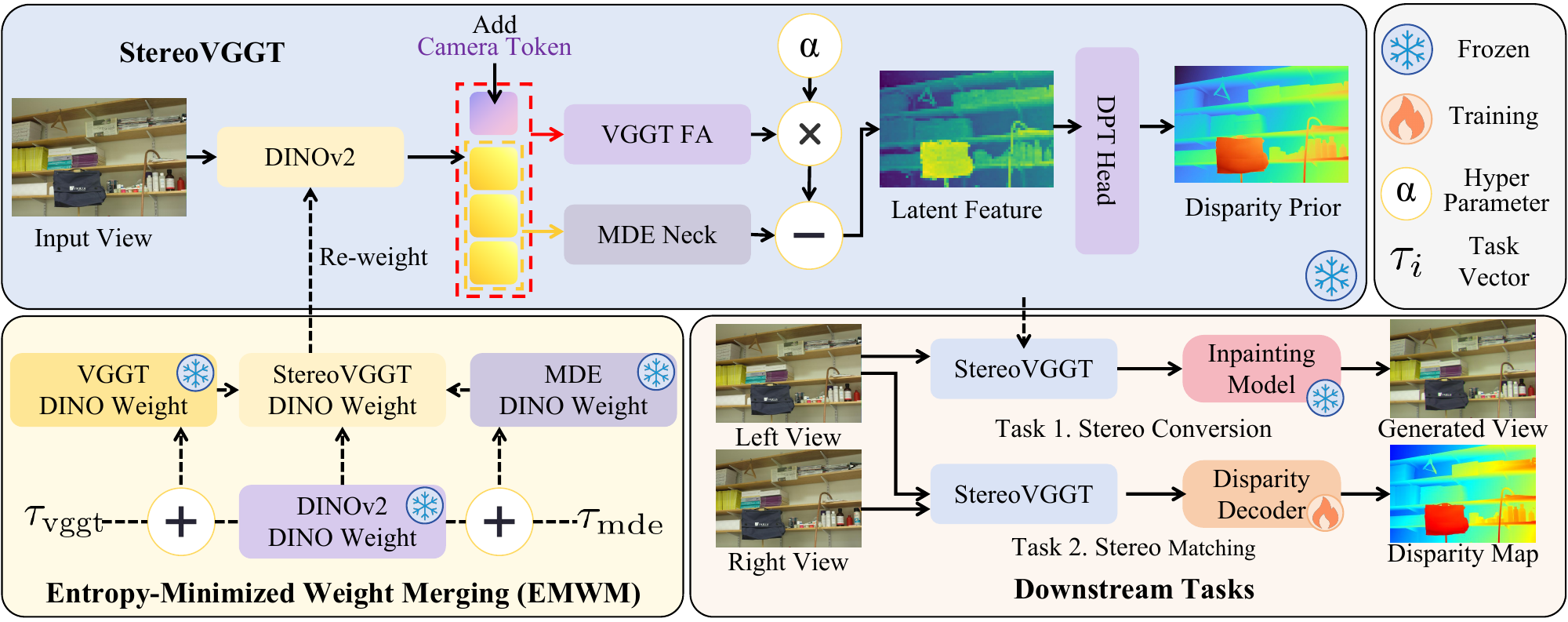} 
	\caption{Overview of the StereoVGGT architecture. StereoVGGT comprises three main stages. First, EMWM synthesizes a new set of optimized DINO weights by merging the weights of VGGT and an MDE model, guided by an entropy-based criterion. Second, the patch tokens generated by the re-weighted DINO are concurrently fed into both frozen VGGT Frame Attention (FA) Blocks and an MDE neck. The VGGT FA features subsequently modulate the MDE neck features, thereby achieving a balance between camera knowledge and fine-grained image-feature representation. Finally, the resulting latent features $X_{stereovggt}$ are passed through a DPT head to generate the disparity prior $d_{stereovggt}$. The latent features and the disparity prior are then leveraged in downstream stereo vision tasks, including stereo conversion and stereo matching.
	} 
	\label{pipeline} 
\end{figure}

\section{Methodology}
\label{sec:method}

\subsection{Overall Framework}
Fig. \ref{pipeline} shows the overall framework of our StereoVGGT and how it works for the downstream applications of stereo vision. 
First, we perform Entropy-Minimized Weight Merging (EMWM) to re-weight the DINO architecture within VGGT, a procedure detailed in Sec. \ref{dm}. 
The resulting re-weighted DINO module serves as the initial component of StereoVGGT. The RGB input $I \in \mathbb{R} ^ {3 \times H \times W}$ 
is processed by this re-weighted DINO module. 
Second, the original feature neck of VGGT is incorporated and integrated with an MDE neck to refine the features, yielding a latent feature $X_{stereovggt}$. This feature extraction process is elaborated upon Sec. \ref{dh}. 
Third, a Dense Prediction Transformer (DPT) head is utilized to compute a disparity prior from the latent feature $X_{stereovggt}$, as detailed in Sec. \ref{dp-head}.
Finally, StereoVGGT can be coupled with various decoders to accommodate different stereo vision downstream tasks. In terms of stereo matching, StereoVGGT is connected to a disparity decoder to estimate the final disparity map, a process elaborated in Sec. \ref{stereo-matching}. 
For the stereo conversion task, StereoVGGT is integrated with an inpainting model to generate the right-side view, as described in Sec. \ref{stereo-conversion}. 

\subsection{Entropy-Minimized Weight Merging (EMWM)}
\label{dm}

\begin{wrapfigure}{r}{0.5\textwidth}
	\centering
	\includegraphics[width=\linewidth]{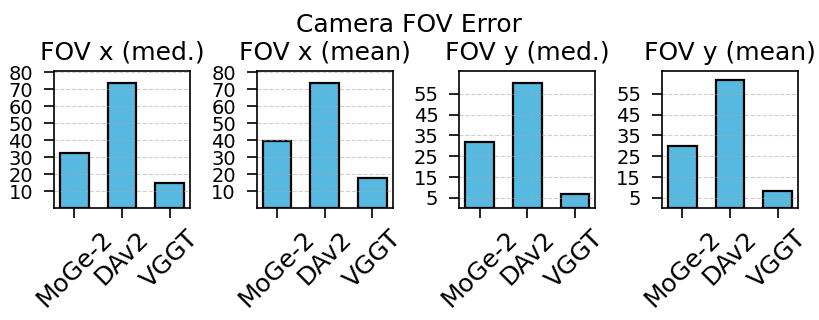} 
	\caption{The bar charts show the median and mean camera FOV errors across different methods. This analysis evaluates FOV estimated from extracted features on the ETH3D dataset. We implement a Levenberg-Marquardt solver to estimate the camera Field-of-View (FOV) \cite{fov} directly from these features. Detailed specifications of this solver are provided in Sec. \ref{solver} 
	} 
	\label{blue} 
\end{wrapfigure}

Figs.~\ref{teaser}--\ref{motivation1} demonstrate that MDE models and 3D foundation models are highly complementary: 3D foundation models encode stronger camera-geometry priors, yet suffer from noticeable structural degradation at the feature level; in contrast, MDE models preserve feature integrity but lack sufficient camera knowledge awareness. We therefore seek an effective way to integrate the strengths of these two types of models.

\begin{figure}[]
	\centering
	\begin{subfigure}[t]{0.49\linewidth}
		\centering
		\includegraphics[width=\linewidth]{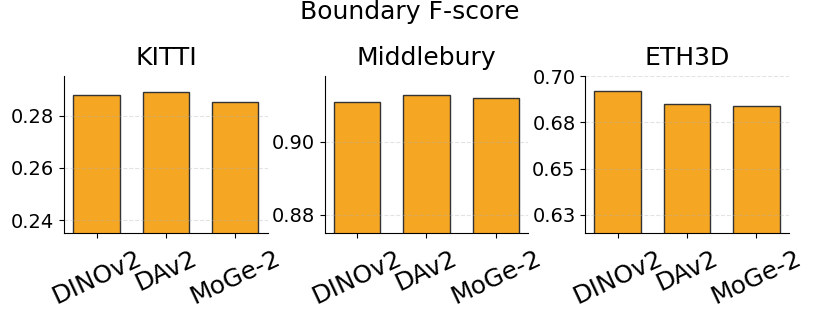}
	\end{subfigure}\hfill
	\begin{subfigure}[t]{0.49\linewidth}
		\centering
		\includegraphics[width=\linewidth]{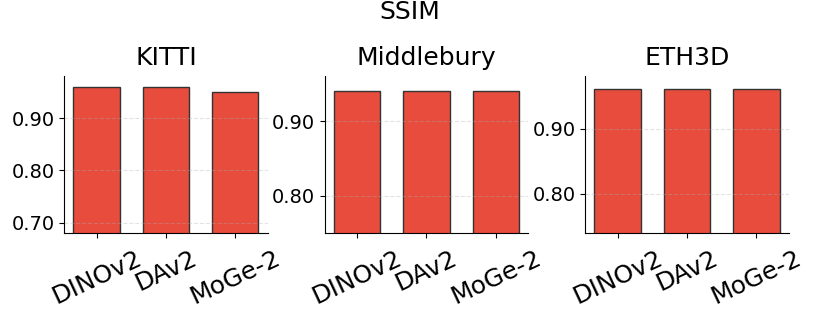}
	\end{subfigure}
	\caption{The orange bar charts represent the Boundary F-score. This metric quantifies the alignment between the boundaries of the feature map and those of the ground-truth disparity map. Concurrently, the red bar charts illustrate the SSIM values calculated between the extracted feature maps and the original input images across the ETH3D, KITTI, and Middlebury datasets. To ensure a fair comparison, all evaluated methods adhere to a consistent evaluation protocol. Comprehensive details are provided in Sec. \ref{ssim}.
	} 
	\label{red} 
\end{figure}

For 3D foundation models, we select VGGT \citep{vggt}. This choice is motivated by the fact that VGGT has been widely adopted as a backbone for a variety of 3D tasks, and that replacing VGGT with larger 3D foundation models has the potential to bring further improvements in many downstream 3D vision tasks. As an early attempt to integrate a 3D foundation model into a stereo vision backbone, our use of VGGT provides a more generally applicable setting and facilitates future studies in adopting our method as a baseline. 
For MDE models, we adopt MoGe-2 \citep{moge} instead of DAv2 \citep{dav2}, which is currently widely used as a stereo vision backbone. Although directly replacing DAv2 with MoGe-2 yields inferior results on the stereo matching task, as shown in Tab. \ref{sceneflow-tab-bridgedepth}, MoGe-2, as a model designed for metric depth estimation, demonstrates stronger camera-awareness than DAv2 and DINOv2, as shown in Fig.~\ref{blue}; moreover, as shown in Fig.~\ref{red}, its feature-preservation capability is comparable to that of DAv2 and DINOv2. We therefore argue that this advantage makes MoGe-2 better suited to handling both stereo matching and stereo conversion tasks within a unified framework. Existing studies have not yet recognized the importance of "activating" camera-awareness for stereo vision tasks, which may explain why current SOTA methods \citep{promptstereo,mono2stereo} predominantly adopt DAv2.

The above discussion explains our rationale for selecting the models integrated by EMWM. The main text provides a high-level overview of EMWM. In the below section, we present the implementation details of EMWM. 

MoGe-2 and VGGT share a common DINO architecture but are trained with different objectives, leading to substantially different weight configurations. EMWM integrates the weights of VGGT and MoGe-2 within their shared DINO architecture. 
Following the definition of task arithmetic \citep{taskvector}, we hypothesize that model parameters can be adapted to a new task by adding a specialized vector $\tau$, referred to as a task vector. We denote the original DINOv2 weights as $\theta$. Note that $\theta$ is not treated as an additional source; instead, it serves as the reference initialization from which task vectors are derived to adapt the model toward specialized domains. 
For the desired resultant weights $\theta_{stereovggt}$, the updated parameter state is governed by
\begin{gather}
	\theta_{stereovggt} = \theta + \tau_{stereovggt},
	\label{eq1}
\end{gather}
where the task vector $\tau_{stereovggt}$ is defined as the weight shift required to adapt the original weights $\theta$ to the specific geometric constraints of stereo vision tasks. 
We posit that the task vector $\tau_{stereovggt}$ can be further formulated as a linear combination of multiple specialized task vectors. This hypothesis allows for the algebraic superposition of distinct capabilities---specifically, the camera-geometry awareness of VGGT and the robust features of MDE models---into a single, unified shift within the parameter space. Formally, this synthesis is expressed as
\begin{gather}
	\tau_{stereovggt} = \lambda_{vggt} \cdot \tau_{vggt} + \lambda_{mde} \cdot \tau_{mde},
	\label{eq2}
\end{gather}
where $\tau_{vggt}$ and $\tau_{mde}$ denote the specialized task vectors derived from VGGT and MoGe-2, respectively. The hyperparameters $\lambda_{mde}$ and $\lambda_{vggt}$ scale their respective task vectors, enabling $\tau_{stereovggt}$ to achieve an effective representation. $\tau_{vggt}=\theta_{vggt}-\theta$ and $\tau_{mde}=\theta_{mde}-\theta$ can be explicitly parameterized by their respective weights, $\theta_{vggt}$ and $\theta_{mde}$, for which $\theta_{stereovggt}  = (1- \lambda_{vggt} - \lambda_{mde} ) \cdot \theta +\lambda_{vggt} \cdot \theta_{vggt} 
+ \lambda_{mde} \cdot \theta_{mde}$ is derived. 
The optimal values of $\lambda_{mde}$ and $\lambda_{vggt}$ are updated iteratively in a data-free manner. Specifically, the merged weights are optimized on a per-layer basis, enabling a flexible merging of the models. For the $l$-th layer of the network architecture, the $i$-th modulated parameter block of StereoVGGT, denoted as $\theta_{stereovggt}^{(l,i)}$, is first converted into a scalar score using its $\ell_2$ norm,
\begin{equation}
	s_i^{(l)} = \left\| \theta_{stereovggt}^{(l,i)} \right\|_2,
\end{equation}
Then mapped onto a probability distribution $P(l)$, which is computed as,
\begin{equation}
	\begin{cases}
		P_i^{(l)} &= \dfrac{\exp(s_i{(l)}/\beta(t))}{\sum_{j=1}^{n}\exp(s_j^{(l)}/\beta(t))},
		\\
		\beta(t) &= \gamma^t
	\end{cases}
\end{equation}
where $t$ signifies the iteration step, and $\beta(t)$ is the temperature at the $t$-th step, $\gamma$ is set to $0.95$ here. The corresponding objective function is given by, 
\begin{equation}
	\min_{\lambda^{(l)}_{vggt},\lambda^{(l)}_{mde}} H^{(l)} = -\sum_{j=1}^{n} P_j^{(l)}\log_2 P_j^{(l)}.
	\label{entropy}
\end{equation}
Following the gradient computation, the parameters $\lambda_{vggt}$ and $\lambda_{mde}$ are updated iteratively according to Eq. (\ref{temp}), yielding intermediate parameters $\lambda^{\prime}_{vggt}$ and $\lambda^{\prime}_{mde}$. 
\begin{gather}
	\begin{cases}
		\lambda_{vggt}^{(l,t+1)\prime} &= \lambda_{vggt}^{(l)} - \eta \cdot \frac{\partial H}{\partial \lambda_{vggt}^{(l)}}, \\
		\lambda_{mde}^{(l,t+1)\prime} &= \lambda_{mde}^{(l)} - \eta \cdot \frac{\partial H}{\partial \lambda_{mde}^{(l)}}.
	\end{cases}
	\label{temp}
\end{gather}
Here, $t$ means the $t$-th iteration. Subsequently, $\lambda^{(l,t+1)\prime}_{vggt}$ and $\lambda^{(l,t+1)\prime}_{mde}$ are projected onto a probability simplex to update parameters $\lambda_{vggt}^{(l,t+1)\prime}$ and $\lambda_{mde}^{(l,t+1)\prime}$, as defined by, 
\begin{equation}
	(\lambda_{vggt}^{(l,t+1)},\lambda_{mde}^{(l,t+1)})=\Pi_{\Delta^1} (\lambda_{vggt}^{(l,t+1)\prime},\lambda_{mde}^{(l,t+1)\prime}).
	\label{simplex}
\end{equation}
\begin{equation}
	\Delta_1=\left\{(\lambda_{vggt}^{(l)},\lambda_{mde}^{(l)}) \;\middle|\; \lambda_{vggt}^{(l)}+\lambda_{mde}^{(l)}=1,\ \lambda_{vggt}^{(l)}\ge 0,\ \lambda_{mde}^{(l)}\ge 0 \right\}.
\end{equation}
It confines the coefficients to a fixed range, enhancing stability to accommodate the training-free nature of the approach. The convergence criterion for $l$-th layer is given by, 
\begin{equation}
	|\lambda_{vggt}^{(l,t+1)} - \lambda_{vggt}^{(l,t)}| \le \varepsilon,~and~ |\lambda_{mde}^{(l,t+1)}- \lambda_{mde}^{(l,t)}| \le \varepsilon,
	\label{converg}
\end{equation}
$\varepsilon=10^{-6}$ here. 
Entropy minimization serves as a data-free criterion for selecting the layer-wise mixing coefficients in EMWM. Our key assumption is that the complementary strengths of VGGT and the MDE model are not uniformly distributed across all parameter blocks: blocks that preserve camera-related geometric priors are expected to be more pronounced in VGGT, whereas blocks that retain fine-grained structural details are expected to be more pronounced in the MDE model. Directly averaging the two task vectors may over-smooth such complementary signals and reduce both camera-awareness and feature fidelity. By minimizing the entropy of the block-norm distribution of the merged weights, EMWM encourages the merged model to concentrate more strongly on a subset of dominant parameter blocks, thereby reducing interference between the two source models. When this optimization is performed in a layer-wise manner, different layers can adaptively favor the source that contributes more useful information at that level, which makes it possible for the merged DINO backbone to preserve structural details while retaining camera-geometry knowledge. We emphasize that this objective is a heuristic proxy rather than a formal guarantee of downstream optimality, but it provides an effective and fully data-free principle for determining the mixing coefficients in practice.

The final DINO weights $\theta_{stereovggt}$ for the $l$-th layer are thus obtained as, 
\begin{equation}
	\theta_{stereovggt}^{(l)}=\lambda_{vggt}^{(l,T)}\theta_{vggt}^{(l)}+\lambda_{mde}^{(l,T)}\theta_{mde}^{(l)}.
	\label{final} 
\end{equation}
The variable $T$ denotes the final iteration count, which is determined by the fulfillment of the termination criteria specified in Eq. (\ref{converg}). Furthermore, a maximum threshold of 20,000 iterations is established; if the convergence conditions are not met within this limit, the iterative process is automatically terminated.

The rationale behind EMWM is that the useful priors contributed by VGGT and the MDE model are likely encoded in different subsets of parameter blocks within their shared DINO backbone. In particular, we hypothesize that VGGT contains blocks more strongly associated with camera-aware geometric reasoning, while the MDE model contains blocks more strongly associated with preserving local structures and fine visual details. If these two sources are merged with poorly chosen coefficients, their task vectors may interfere with each other, producing an overly averaged parameterization that weakens both types of useful signals. To mitigate this issue without relying on training data, we measure the block-wise energy distribution of the merged weights and optimize the layer-wise coefficients to minimize its entropy. This objective favors a sharper distribution over parameter blocks, meaning that the merged model is encouraged to retain a more selective set of dominant blocks instead of diffusing weight mass uniformly across all blocks. Under our assumption that the dominant blocks are more likely to carry useful geometric or structure-preserving priors, entropy minimization reduces fusion interference and leads to a more effective combination of the two models. Since the optimization is performed independently for each layer, the resulting coefficients can adapt to different functional roles across the network, allowing some layers to lean more toward structural preservation and others toward camera knowledge. Importantly, this objective should be interpreted as a practical proxy rather than a theoretical proof that the resulting coefficients are globally optimal for downstream stereo tasks; its value lies in providing a stable, interpretable, and fully data-free mechanism for weight merging.

\subsection{Feature Neck}
\label{dh}
To further balance feature quality and camera knowledge, we propose a dual-branch design in which VGGT neck features modulate MDE neck features. 
This neck architecture is formulated as $X_{FA} = N_{FA}(F_{stereovggt}, ct)$, where $F_{stereovggt}$ denotes the output of the re-weighted DINO module.
Camera token $ct$ is integrated via VGGT FA $N_{FA}$ with the patch tokens from the StereoVGGT re-weighted DINO, enriching the pipeline with camera geometry knowledge. 
We intentionally excluded the Global Attention (GA) component of the Alternative Attention (AA) Block in VGGT neck. Since StereoVGGT uses only a single view as input, applying both global and frame attention is functionally redundant. 
These enriched patch tokens, $X_{FA}$, are employed to modulate the features derived from the MDE neck, $N_{MDENeck}$, thereby generating the final output representation $X_{stereovggt}$, 
\begin{equation}
	X_{stereovggt} = N_{MDENeck}(F_{stereovggt}) - \alpha \cdot X_{FA}
	\label{neck}
\end{equation}
Here, 
$\alpha=0.2$ is a hyperparameter to modulate these features. 
We employ feature-wise subtraction to integrate the two branches. Given that the two feature tensors are defined in a comparable latent space with identical spatial and channel dimensions, subtraction provides a simple residual interaction that highlights complementary information between the branches rather than merely aggregating their responses.

\subsection{Disparity Prior Head}
\label{dp-head}
The frozen Dense Prediction Transformer ($DPT$) \citep{dpt} depth head within the original VGGT first predicts depth, which is then converted into a disparity prior $d_{stereovggt}$ and can be expressed as, 
\begin{equation}
	\hat{z}=DPT(X_{stereovggt}),\quad d_{stereovggt} = \frac{f \cdot B}{\hat{z}}
	\label{dpt}
\end{equation}
where $f$ represents the focal length, $B$ is the baseline length. 
In the stereo matching task, $f$ is known, whereas in the stereo conversion task, it is estimated by the frozen VGGT camera head. The $B$ is known for both two stereo vision tasks. 

\subsection{Stereo Matching Disparity Decoder}
\label{stereo-matching}

For the stereo matching task, we adopt IGEV-Stereo \citep{igev} as the baseline framework. This choice is motivated by the widespread use of IGEV-Stereo as a benchmark architecture for evaluating the effectiveness of different feature backbones \citep{mocha,igev++,gip}. By leveraging its well-established stereo matching pipeline, we are able to isolate and systematically assess the contribution of the proposed StereoVGGT representations. 

Adhering to the architectural paradigm of IGEV-Stereo \citep{igev}, the feature representations $X_{stereovggt}^{I_l}$ and $X_{stereovggt}^{I_r}$ are first used to construct an all-pairs correlation volume that encodes dense matching similarities across the stereo pair. This correlation volume is further integrated with a Geometry Encoding Volume, yielding a Combined Geometry Encoding Volume (CGEV). 
Unlike conventional correlation volumes that only model pairwise feature similarity, the CGEV explicitly embeds geometricrelationships through lightweight 3D aggregation, providing a more informative representation for disparity estimation.

Based on the constructed CGEV, an initial disparity estimate
$d_0$ is first regressed. This coarse prediction serves as a starting point for iterative optimization and significantly accelerates
convergence compared with purely recurrent stereo architectures.
Subsequently, a recurrent Update Operator implemented using ConvGRUs \citep{raft,igev} iteratively refines the disparity field. At each iteration, the current disparity hypothesis indexes the CGEV to
retrieve geometry-aware matching features, which are then fused with the hidden recurrent state to predict disparity updates. Formally, the refinement process can be expressed as,
\begin{equation}
	d^{t+1}=d^{t}+\Delta d^{t},
\end{equation}
where $\Delta d^{t}$ denotes the disparity residual predicted by the
ConvGRU update block at iteration $t$.

The entire stereo matching pipeline is encapsulated in a disparity decoder $\phi_{decoder}$, which maps the StereoVGGT representations directly to the final disparity prediction,
\begin{equation}
	d = \phi_{decoder}(X_{stereovggt}^{I_l},X_{stereovggt}^{I_r}),
	\label{sm}
\end{equation}
where $d$ denotes the final refined disparity map. Notably, the decoder architecture remains identical to that of IGEV-Stereo, ensuring that any performance gain can be attributed solely to the representations learned by StereoVGGT rather than modifications to the stereo matching decoder.

\subsection{Stereo Conversion Inpainting Model}
\label{stereo-conversion}
To demonstrate the potential of StereoVGGT for stereo conversion, Mono2Stereo is used as our baseline framework. 
Mono2Stereo was chosen because it proposes two standardized benchmark datasets for stereo conversion, a domain where established quantitative benchmarks are generally lacking. 
Consistent with other leading methods, Mono2Stereo adopts a two-stage paradigm: it utilizes an MDE model to extract disparity prior in the first stage, followed by an inpainting model $\phi_{inpainting}$ to synthesize the right-view image $I_{r~(generated)}$ in the second stage. We replace the original DAv2 within the Mono2Stereo framework with our proposed StereoVGGT for the first stage.  Following the architecture of Mono2Stereo, we employ the Marigold VAE \citep{marigold}---pre-trained on the Mono2Stereo datasets---as the inpainting model for synthesized view completion. Specifically, the disparity prior $d_{stereovggt}$ produced by StereoVGGT, as defined in Eq. (\ref{dpt}), is fed into the inpainting model $\phi_{inpainting}$ alongside the left-view image $I_l$ to synthesize the generated right-view image $I_{r~(generated)}$. This process is formally expressed as

\begin{gather}
I_{r~(generated)} = \phi_{inpainting}(d_{stereovggt}, I_l )
\label{sc1}
\end{gather}

The weights for both the StereoVGGT and inpainting modules are kept frozen, enabling the entire process to be training-free.

The weights for both the StereoVGGT and inpainting modules are kept frozen, enabling the entire process to be training-free.

\begin{table}[t]
	\centering
	\setlength{\tabcolsep}{1.0mm}
	\begin{tabular}{c|c|c|ccccc|ccc}
		\toprule[1.5 pt]
		\multirow{2}{*}{Feature Backbone} & \multirow{2}{*}{Disparity Decoder} &\multirow{2}{*}{Reference}          & & \multicolumn{3}{c}{Non-occ $\downarrow$} &  & \multicolumn{3}{c}{All $\downarrow$} \\ 
		&   &  &  & all     & fg      & bg      &  & all    & fg     & bg    \\ \midrule
		ViT-Large &  ViT-Base+DPT  &CroCo-Stereo &   & 1.51    & 2.56    & 1.30    &  & 1.59   & 2.65   & 1.38  \\
		MobileNet V2&  CGEV+SRU   &Selective-IGEV  &   & 1.44    & 2.55    & 1.22    &  & 1.55   & 2.61   & 1.33  \\
		DAv2  &   SGA\&MGR+ConvGRU    & MonSter  &  & 1.33    & 2.76    & \cellcolor{orange!20}1.05    &  & \cellcolor{red!20}1.41  & 2.81   & \cellcolor{orange!20}1.13  \\ 
		DAv2   &   SGA\&MGR+PRU   & PromptStereo &  & \cellcolor{orange!20}1.32    & 2.76    & \cellcolor{red!20}1.04    &  & \cellcolor{red!20}1.41  & 2.85   & \cellcolor{red!20}1.12  \\ 
		\midrule \midrule
		MobileNet V2 & \multirow{10}{*}{\begin{tabular}[c]{@{}c@{}} CGEV+ConvGRU\\Paradigm\\ (IGEV-Stereo) \end{tabular}}   &IGEV-Stereo  &   & 1.49    & 2.62    & 1.27    &  & 1.59   & 2.67   & 1.38  \\
		EfficientNet V2  &  & MoCha-Stereo &  & 1.44    & \cellcolor{orange!20}2.42   & 1.24    &  & 1.53   & \cellcolor{orange!20}2.43   & 1.36  \\
		MiDaS  &  &  LoS  & & 1.52 & 2.66    & 1.29   &  & 1.65   & 2.81   & 1.42  \\
		ConvNeXt V2 &    &GIP-Stereo       &   & 1.47    & 2.64    & 1.20   &  & 1.51   & 2.71   & 1.27  \\
		Depth Anything & & Mono2Stereo  &  & 1.48    & 2.59   & 1.26    &  & 1.58   & 2.67   & 1.36  \\
		DINOv2 et al.   &  &ViTAStereo&   & 1.41    & 2.90    & 1.12   &  & 1.50  & 2.99   & 1.21  \\
		DAv2   &  & IGEV++  & &  \cellcolor{green!10}1.36 &2.80 & \cellcolor{green!10}1.07 & & \cellcolor{green!10}1.43& 2.80& \cellcolor{green!10}1.15 \\
		DAv2 et al.& &AIO-Stereo &  &
		1.43 & \cellcolor{green!10}2.51& 1.22 && 1.54& \cellcolor{green!10}2.57& 1.34 \\
		VGGT &   &  - &   & 1.58 & 2.87 & 1.33 &  & 1.67 & 2.81 & 1.44 \\
		\rowcolor{yellow!20}
		Ours& & - & & \cellcolor{red!20}1.31 & \cellcolor{red!20}2.31 & 1.12 && \cellcolor{orange!20}1.42 & \cellcolor{red!20}2.38&  1.22 \\
		\bottomrule[1.5 pt]    
	\end{tabular}
	\caption{Evaluation results on the KITTI benchmark under 3-pixel error threshold. "Non-occ" and "All" denote evaluation on non-occluded pixels and all pixels, respectively. "all" indicates the outlier percentage over all ground-truth pixels, "fg" and "bg" correspond to foreground and background regions. 
	Red, orange, and green shading indicate \colorbox{red!20}{1st}, \colorbox{orange!20}{2nd}, and \colorbox{green!10}{3rd} place, respectively. 
	The row highlighted with yellow denotes \colorbox{yellow!20}{our proposed method}. 
	}
	\label{kitti15}
\end{table}

\section{Experiments}
\label{sec:exp}

This section provides a comprehensive evaluation of the proposed StereoVGGT's effectiveness across various stereo vision tasks. To validate the efficacy of StereoVGGT, we conduct two sets of experiments:
\begin{itemize}
	\item Experiments were conducted on the KITTI and Scene Flow datasets to evaluate the effectiveness of StereoVGGT for stereo matching (Sec. \ref{stereo-matching-exp}).
	\item Experiments were conducted on the Mono2Stereo and Inria 3D Movie datasets to evaluate the performance of StereoVGGT for stereo conversion (Sec. \ref{stereo-conversion-exp}).
	\item Ablation studies were conducted on the KITTI, ETH3D, and Middlebury datasets to evaluate the investigate the individual contributions of StereoVGGT component for monocular depth estimation. (Sec. \ref{ablations})
\end{itemize}

\subsection{Implementation Details}

Our implementation is built upon PyTorch and runs in a CUDA-enabled environment for GPU acceleration. All experiments are conducted on NVIDIA RTX 3090 GPUs. 
All evaluations are performed using mixed-precision inference to maximize computational efficiency while preserving numerical stability. For fair comparison, the same hardware and software configuration is adopted across all competing methods. 
The architecture of StereoVGGT consists of a VGGT DINO image encoder (VGGT branch), a MoGe-2 DINO branch (MDE branch), and the frame-attention aggregator inherited from VGGT. Following DINOv2, intermediate features from layers ${5,11,17,23}$ are extracted and projected to a unified 1024-dimensional representation using $1\times1$ convolutions. The projected features are aggregated into a shared token representation, to which the pretrained VGGT camera token and register tokens are appended. The resulting sequence is subsequently processed by 24 pretrained VGGT frame-attention blocks. All experiments are conducted with the random seed set to 1234.

\subsection{Stereo Matching}
\label{stereo-matching-exp}
To evaluate the potential of StereoVGGT for stereo matching, we build a pipeline consisting of the EMWM, the Feature Neck and the Stereo Matching Disparity Decoder, as illustrated in Sec. \ref{stereo-matching}.

\begin{figure*}[tp]
	\centering
	\includegraphics[width=1.0\linewidth]{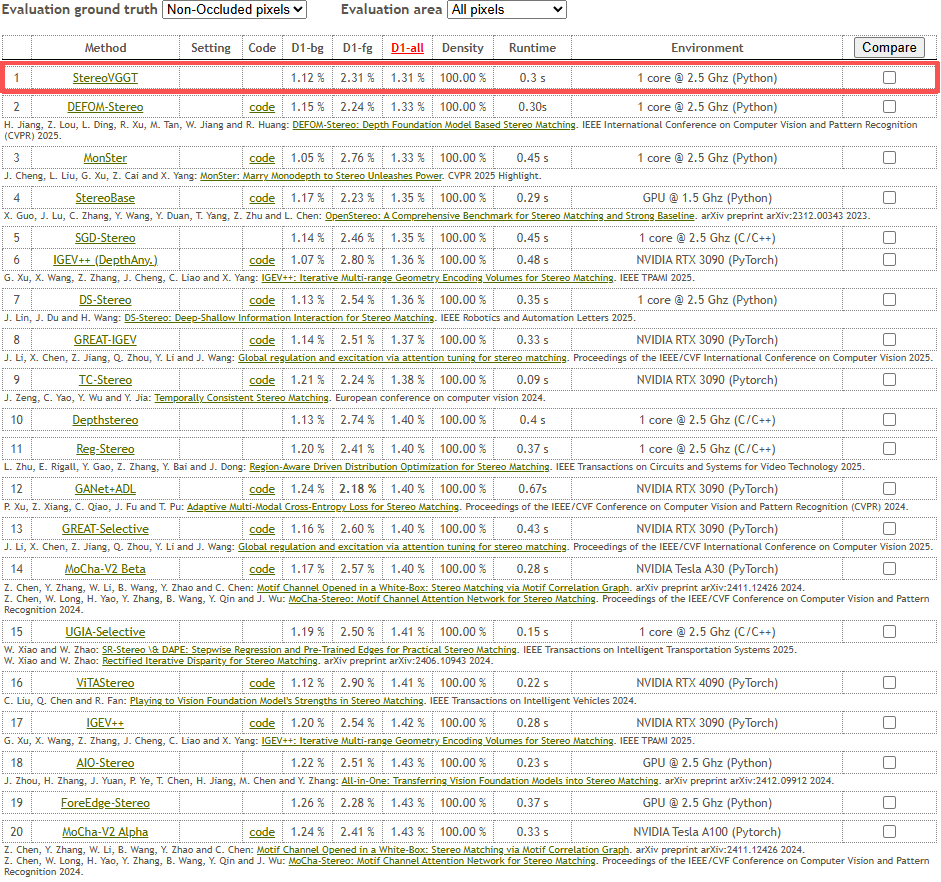} 
	\caption{Our stereo matching network built upon StereoVGGT ranked $1^{st}$ on the KITTI benchmark for non-occluded pixels at the time of submission.} 
	\label{snap} 
\end{figure*}

\begin{figure}[t]
	\centering
	\includegraphics[width=\linewidth]{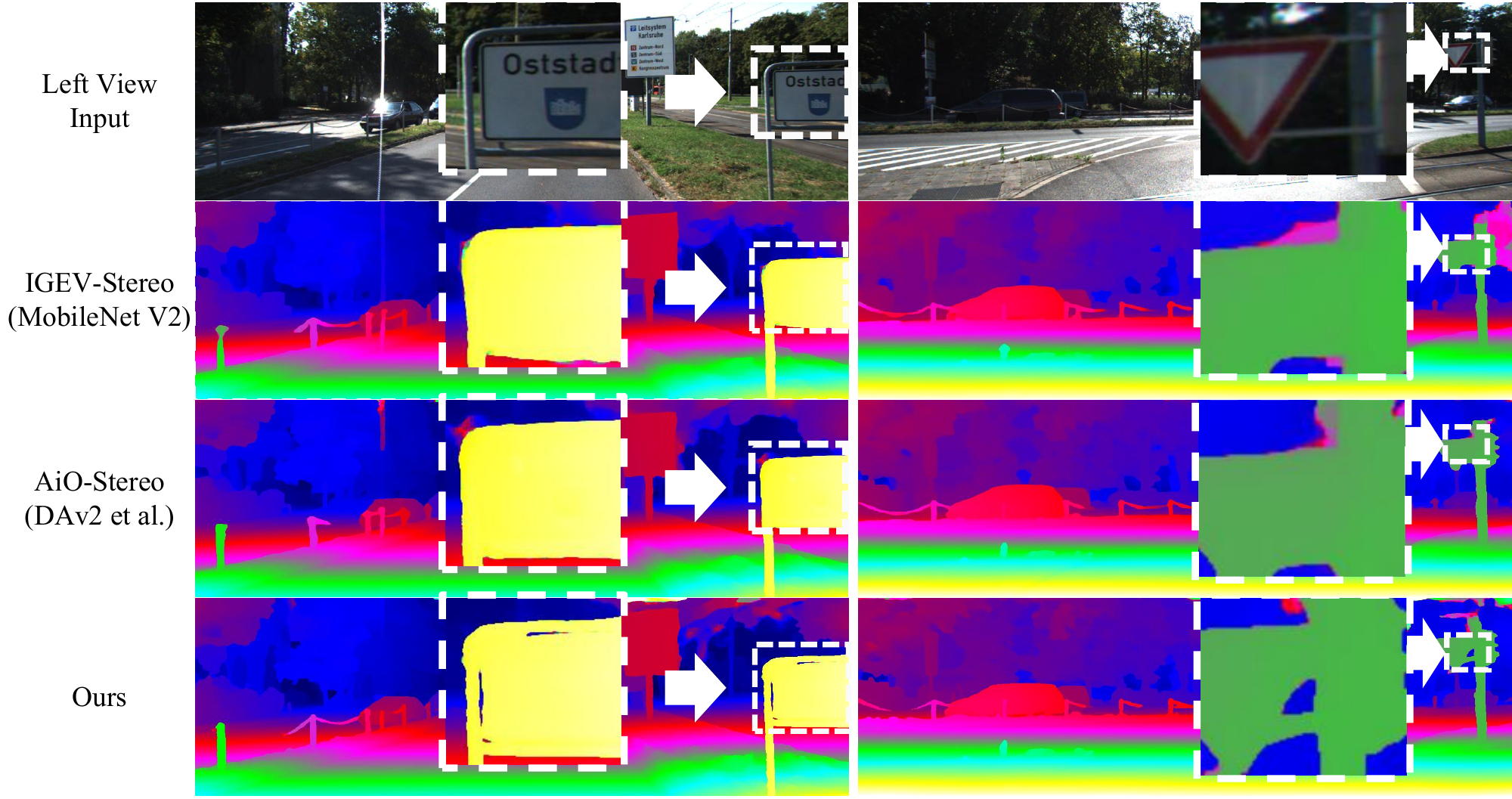} 
	\caption{Visualization comparison on the KITTI dataset. The text in parentheses indicates the feature backbone used by each method. 
	The preceding examples show that both input scenes contain gaps between the signs and poles. } 
	\label{vis-kitti} 
\end{figure}

\subsubsection{KITTI} 
The lower half of Tab. \ref{kitti15} compares different feature backbones integrated with the same IGEV-Stereo Disparity Decoder framework. They are all trained on the Scene Flow and KITTI training sets. 
The results indicate that employing the proposed StereoVGGT as the backbone yields superior performance gains over backbones derived from MDE models \citep{mono-stereo,aio} and PVMs \citep{gip,aio}. 
In comparison to the MobileNet V2 baseline, StereoVGGT reduces the error rates by 12.08\%, 11.83\%, 11.81\%, 10.69\%, 10.86\%, and 11.59\% on the six metrics, respectively. 
The upper part of Tab. \ref{kitti15} compares StereoVGGT with other SOTA methods. 
Our stereo matching network built upon StereoVGGT ranked $1^{st}$ on the KITTI benchmark for non-occluded pixels at the time of submission, as shown in Fig. \ref{snap}. 
The effectiveness of our method is further corroborated by the qualitative visualizations provided in Fig. \ref{vis-kitti}. A key difference is that IGEV-Stereo and AiO-Stereo fail to reconstruct these holes, whereas our method successfully recognizes them as part of the distant background.

\subsubsection{Scene Flow} 

A quantitative comparison of StereoVGGT with competing approaches on the Scene Flow dataset is provided in the upper part of Tab.~\ref{sceneflow-tab-bridgedepth}. All experiments are trained exclusively on the Scene Flow training set. 
In addition to replacing the MobileNetV2 baseline with StereoVGGT, we conducted a backbone replacement comparison by substituting the backbone with several other advanced architectures, including EfficientNet V2, VGGT, MoGe-2, and DAv2. 
Experimental results demonstrate that StereoVGGT consistently outperforms SOTA backbone variants across the evaluated metrics.

To further evaluate the generalization of StereoVGGT, we conducted backbone substitution experiments within the BridgeDepth framework \citep{bridgedepth}, as shown in the lower part of Tab. \ref{sceneflow-tab-bridgedepth}. BridgeDepth, a SOTA model for the Scene Flow dataset \citep{sceneflow}, is designed to support the seamless integration of frozen monocular depth estimation models. In our experiments, we replaced the default DAv2 \citep{dav2} backbone with MoGe-2 \citep{moge}, VGGT \citep{vggt}, and our proposed StereoVGGT. The empirical results demonstrate that StereoVGGT consistently remains the most effective variant among the evaluated configurations, further substantiating its robustness across different stereo matching pipelines.

\begin{table*}[t]
	\centering
	\setlength{\tabcolsep}{1.0mm}
	\begin{tabular}{c|c|c|cc}
		\toprule[1.5 pt]
		Baseline Network  &Type                  & Backbone & D1 $\downarrow$      & EPE $\downarrow$     \\ \midrule
		\multirow{6}{*}{IGEV-Stereo \citep{igev}} & \multirow{2}{*}{PVM} & \cellcolor{gray!10}MobileNet V2  &  \cellcolor{gray!10}5.3  &  \cellcolor{gray!10}0.47        \\
		& &EfficientNet V2  $\dagger$  &    5.2 &  0.46      \\
		& 3R&VGGT  $\dagger$   &  5.3   &  0.46  \\
		& \multirow{2}{*}{MDE}&MoGe-2  $\dagger$  &  5.1   &  0.45  \\
		& &DAv2 $\dagger$  &    5.1 &  0.44    \\
		& \multicolumn{2}{c|}{\cellcolor{yellow!20} Ours}  & \cellcolor{yellow!20}4.9    &  \cellcolor{yellow!20}0.43    \\ \midrule \multirow{4}{*}{BridgeDepth \citep{bridgedepth}} & \multirow{2}{*}{MDE}   & \cellcolor{gray!10}DAv2   & \cellcolor{orange!20}3.7 & \cellcolor{orange!20}0.37       \\
		& & MoGe-2  $\dagger$ &  \cellcolor{orange!20}3.7   &  \cellcolor{green!10}0.39  \\
		&3R & VGGT  $\dagger$ &   \cellcolor{green!10}4.0  &   0.42 \\
		&\multicolumn{2}{c|}{\cellcolor{yellow!20} Ours}  & \cellcolor{red!20}3.3    &  \cellcolor{red!20}0.33     \\
		\bottomrule[1.5 pt]    
	\end{tabular}
	\caption{Quantitative evaluation on Scene Flow test set. The error threshold is 1 px.  $ \dagger $ denotes our reproduced version, for which no publicly available source code exists. Light Gray shading indicates the \colorbox{gray!10}{baseline method}. The row highlighted with yellow denotes \colorbox{yellow!20}{our proposed method}. }
	\label{sceneflow-tab-bridgedepth}
\end{table*}
\begin{table}[h]
	\centering
	\setlength{\tabcolsep}{1.0mm}
	\begin{tabular}{cc|c|cccc}
		\toprule[1.5 pt]
		Disparity Stage   & Inpainting Stage& Reference  & RMSE $\downarrow$ & SSIM $\uparrow$& SIoU $\uparrow$ & PSNR $\uparrow$ \\ \midrule
		DPT-based
		&SD2-depth &   StereoDiffusion&      9.345  & 0.2703    & 0.1941 & 28.75   \\
		MiDaS 
		+Stereo Pixel Shift           &    Stable Diffusion 
		&StereoDiffusion& 8.451   & 0.4472    & 0.2672 & 29.66   \\
		DepthCrafter     & SVD &   StereoCrafter&      7.392    &  0.5866      &     0.2656     &   29.71     \\ 
		\midrule  \midrule
		DAv2  & \multirow{4}{*}{\begin{tabular}[c]{@{}c@{}}Marigold VAE \\(Fine-tuned by\\Mono2Stereo) \end{tabular}}& Mono2Stereo & \cellcolor{green!10}6.857 &  \cellcolor{green!10}0.7227 &  \cellcolor{orange!20}0.2859  & \cellcolor{green!10}31.53 \\   \cmidrule{1-1} \cmidrule{3-7}           
		VGGT &  &  -&  \cellcolor{orange!20}6.734 &       0.7082 &  \cellcolor{green!10}0.2801  &  \cellcolor{orange!20}31.57    \\
		MoGe-2  &    &  -&   6.891 & \cellcolor{orange!20}0.7247 &  0.2753  &   30.88   \\
		\rowcolor{yellow!20} Ours & & - & \cellcolor{red!20}6.462&  \cellcolor{red!20}0.7343 & \cellcolor{red!20}0.2952  &   \cellcolor{red!20}32.03    \\
		\bottomrule[1.5 pt]    
	\end{tabular}
	\caption{Stereo conversion evaluation on the Inria 3D Movie dataset. In the upper table, StereoDiffusion uses SD2-depth as its inpainting model, while StereoCrafter uses SVD. In the lower table, both methods employ the Marigold VAE fine-tuned by Mono2Stereo as their inpainting model.
	}
	\label{inria}
\end{table}
\begin{figure*}[t]
	\centering
	\centering
	\includegraphics[width=\linewidth]{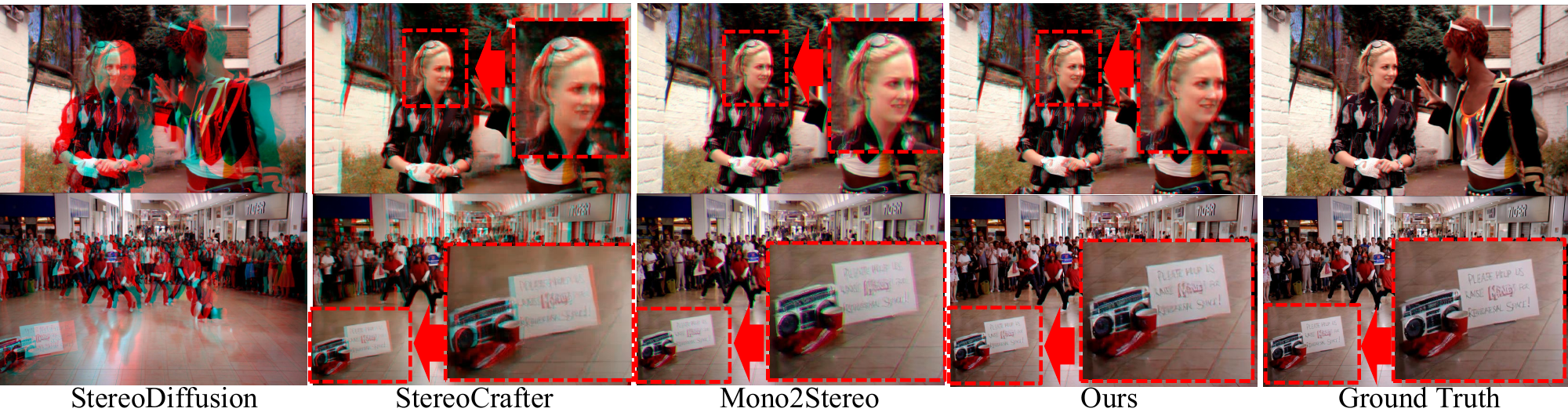} 
	\caption{Visual comparison in anaglyph 3D format on the Inria 3D Movie dataset. The red boxes highlight the main differences.} 
	\label{vis-anaglyph} 
\end{figure*}

\begin{table}[t]
	\centering
	\setlength{\tabcolsep}{0.5mm}
	\begin{tabular}{c|c|cccc}
		\toprule[1.5 pt]
		Scenes&  Method & RMSE $\downarrow$& SSIM $\uparrow$ &SIoU $\uparrow$  & PSNR $\uparrow$\\ \midrule
		\multirow{2}{*}{Indoor}    & Mono2Stereo&   5.218  &    0.8161  &  0.2661   &   34.12    \\
		& Ours &   \textbf{5.192}&   \textbf{0.8354}  & \textbf{0.2689}       &  \textbf{34.20}     \\ \midrule
		\multirow{2}{*}{Outdoor}   & Mono2Stereo&  5.655  &   0.7653 &  0.2708    &   33.51     \\
		& Ours&   \textbf{5.553} & \textbf{0.7970} & \textbf{0.2743}    & \textbf{33.72}   \\ \midrule
		\multirow{2}{*}{Animation}  & Mono2Stereo&   5.763&   0.7628 &  0.2755    &   33.13     \\
		& Ours&  \textbf{5.716}&  \textbf{0.7904}  & \textbf{0.2798}     &   \textbf{33.19}     \\ \midrule
		\multirow{2}{*}{Complex}    & Mono2Stereo&   5.786&   0.7810&  0.2639     &   \textbf{33.59}     \\
		& Ours&   \textbf{5.626}&   \textbf{0.8018}   &   \textbf{0.2663}     &   33.49    \\ \midrule
		\multirow{2}{*}{Simple}    & Mono2Stereo&    4.281 &   0.8507    &   \textbf{0.2818}      &   36.08      \\
		& Ours&   \textbf{4.276}&   \textbf{0.8804} &   0.2815       &   \textbf{36.09}     \\ 
		\bottomrule[1.5 pt]    
	\end{tabular}
	\caption{Quantitative comparison of our proposed method and Mono2Stereo on the Mono2Stereo dataset. Bold font indicates the \textbf{better-performing} result.}
	\label{mono2stereo}
\end{table}
\begin{table}[t]
	\centering
	\begin{tabular}{c|ccc|cc}
		\toprule[1.5 pt]
		& \multicolumn{3}{c|}{Stereo Matching} & \multicolumn{2}{c}{Stereo Conversion} \\
		& Middlebury       & KITTI      & ETH3D      & Inria 3D         & Mono2Stereo         \\ 
		& (Bad-2$\downarrow$)    & (Bad-3$\downarrow$)   & (Bad-1$\downarrow$)  & \multicolumn{2}{c}{(RMSE$\downarrow$)}     \\ 
		\midrule
		Baseline      & 6.21            & 6.03      & 4.05   & 6.86                   & 5.34                \\ \midrule
		Ours    &         \textbf{4.89}         &       \textbf{5.52}     &     \textbf{3.37}       & \textbf{6.46}                   & \textbf{4.14}                \\
		w/o MDE DINO   &          6.20        &     6.04       &     4.01       &     6.58             &         4.36            \\
		w/o VGGT DINO &          5.85        &       5.89     &      3.97      &           6.62             &         4.51            \\
		w/o MDE Neck   &      5.21     &      5.83      &      3.65     &     6.49                   &           4.23          \\
		w/o VGGT FA  &        5.00       &    5.61    &    3.48      &           6.51             &           4.28         \\
		\bottomrule[1.5 pt]    
	\end{tabular}
	\caption{Ablation studies. The results for stereo matching are obtained via zero-shot inference using weights trained on Scene Flow. The 2px error rate is employed for Middlebury, 3px error rate for KITTI, and 1px error rate for ETH3D. "w/o MDE DINO" denotes EMWM re-weighted only by VGGT, "w/o VGGT DINO" denotes EMWM re-weighted only by MoGe-2.}
	\label{abl}
\end{table}

\begin{table*}[t]
	\centering
	\setlength{\tabcolsep}{0.5mm}
	\begin{tabular}{c|c|cccccc|cccccc|ccccc}
		\toprule[1.5 pt]
		\multirow{2}{*}{Type}&\multirow{2}{*}{Method}&  & \multicolumn{4}{c}{KITTI $\downarrow$}     &  &  & \multicolumn{4}{c}{ETH3D $\downarrow$}    &  &  & \multicolumn{4}{c}{Middlebury $\downarrow$}  \\ 
		&  &  & EPE   & 7 px  & 9 px  & 15 px &  &  & EPE  & 7 px  & 9 px  & 15 px &  &  & EPE    & 7 px  & 9 px  & 15 px \\  \midrule
		Gen.  &ZeroStereo &  & 27.91 & 77.48 & 72.40 & 58.85 &  &  &  23.85    &   68.35    &   61.71    &   48.66    &  &  &    19.98    &   80.43    &    72.61   &  52.81   \\  
		\midrule
		\multirow{3}{*}{MDE}&Unidepth V2    &  &   9.07    &   63.64    &   47.74    &   9.97    &  &  &   \cellcolor{orange!20}3.03   &    \cellcolor{orange!20}15.06   &   \cellcolor{orange!20}9.13    &    \cellcolor{orange!20}0.21   &  &  &     27.72   &   89.99    &   84.13    & 70.17  \\
		&MoGe-2        &  &   \cellcolor{green!10}6.74    &   \cellcolor{orange!20}32.16   &   \cellcolor{orange!20}18.58    &    \cellcolor{orange!20}6.15   &  &  &   \cellcolor{green!10}4.84   &   \cellcolor{green!10}19.97    &    \cellcolor{green!10}12.01   &    \cellcolor{green!10}0.30   &  &  &   \cellcolor{orange!20}18.11     &   \cellcolor{orange!20}68.80    &    \cellcolor{orange!20}52.14   &  \cellcolor{orange!20}39.57   \\  
		&DAv2  &  & \cellcolor{orange!20}5.88  & \cellcolor{green!10}32.91 & \cellcolor{green!10}23.21 & \cellcolor{green!10}7.19  &  &  & 8.24 & 36.17 & 28.17 & 16.70 &  &  & 24.94  & 86.90 & 82.91 & 67.44\\
		\midrule
		\multirow{2}{*}{3R}&VGGT       &  & 13.07 & 68.40 & 58.06 & 33.99 &  &  &   13.11   &    69.88   &  54.31     &   28.21    &  &  &     18.71   &    71.25   &    60.99   &  \cellcolor{green!10}  40.87   \\
		&fastVGGT      &  &    13.07   &   68.41    &   58.11    &   34.05    &  &  &   13.11   &   69.89    &    54.32   &    28.21   &  &  &    \cellcolor{green!10}18.70    &    \cellcolor{green!10}70.77   &  \cellcolor{green!10} 60.44    &    41.01  \\  \midrule
		\multicolumn{2}{c|}{Ours}   &  & \cellcolor{red!20}2.71  & \cellcolor{red!20}6.87  & \cellcolor{red!20}3.85  & \cellcolor{red!20}1.03  &  &  & \cellcolor{red!20}2.38 & \cellcolor{red!20}10.70 & \cellcolor{red!20}7.27  & \cellcolor{red!20}0.16  &  &  & \cellcolor{red!20}15.14  & \cellcolor{red!20}63.71 & \cellcolor{red!20}52.04 & \cellcolor{red!20}35.21 \\
		\bottomrule[1.5 pt]
	\end{tabular}
	\caption{Cross-dataset performance comparison of disparity estimation in a monocular setting on binocular stereo datasets. The input consists of the left view of a stereo pair, along with the camera focal and baseline length. All models were evaluated in their pre-trained state without any fine-tuning. 
	"Gen.": Generation-based method, "MDE": Monocular Depth Estimation method, "3R": 3D Reconstruction method. 
	}
	\label{mde}
\end{table*}

\subsection{Stereo Conversion}
\label{stereo-conversion-exp}
Following the methodology description, we developed a pipeline to assess the efficacy of StereoVGGT for stereo conversion.

\subsubsection{Inria 3D Movie}

As depicted in Tab. \ref{inria}, our stereo conversion method achieves SOTA performance across all metrics. The visual comparison in Fig. \ref{vis-anaglyph} demonstrates that our method produces anaglyph 3D scenes with superior clarity.

\subsubsection{Mono2Stereo} 
The Mono2Stereo method was specifically trained on this dataset. Within the Mono2Stereo framework, we replaced its original DAv2 disparity stage with StereoVGGT and compared the resulting performance. 
The results in Tab.~\ref{mono2stereo} indicate that the StereoVGGT-based version of the Mono2Stereo pipeline achieves better overall performance in all five scenarios and outperforms the original pipeline on 18 of the 20 metrics.

\subsection{Ablation Study}
\subsubsection{Downstream Stereo Vision Tasks}
Ablation studies are conducted on each module across two tasks and five datasets, as shown in Tab. \ref{abl}. 
The baseline for stereo matching is IGEV-Stereo \citep{igev} with MobileNetV2 \citep{mobilenetv2} as the backbone, whereas the baseline for stereo conversion is Mono2Stereo \citep{mono2stereo} with DAv2 \citep{dav2} as the backbone. "w/o MDE DINO" denotes the performance obtained without EMWM-based DINO weight fusion, using only the VGGT DINO weights. "w/o VGGT DINO" denotes the performance obtained without EMWM-based DINO weight fusion, using only the MoGe-2 \citep{moge} DINO weights. "w/o MDE Neck" denotes the performance obtained when the feature neck relies only on VGGT FA. "w/o VGGT FA" denotes the performance obtained when the feature neck relies only on the MDE neck. The results show that removing any module leads to performance degradation in both stereo matching and stereo conversion.

\subsubsection{Monocular Depth Estimation Using Only StereoVGGT}
\label{ablations}

To further evaluate the contribution of individual StereoVGGT components to stereo vision features, we formulate the following research questions and conduct corresponding ablation studies on standard binocular stereo matching datasets, including KITTI \citep{kitti}, Middlebury \citep{middlebury}, and ETH3D \citep{eth3d}. We leverage the properties described in Eq. (\ref{dpt})  to evaluate the dataset using the disparity estimated under a monocular setting. 
In contrast to the settings in the Experiments section of the main paper, in this section, we utilize only the left-view image and camera intrinsics (focal length $f$ and baseline length $B$) as inputs to generate the predicted disparity map.
To avoid introducing an inherent bias toward models that directly predict metric depth, all competing methods are evaluated using an optimally scaled prediction. Specifically, a global scale factor is selected for each model to minimize the resulting disparity error, thereby providing a favorable upper-bound estimate of its monocular disparity performance. This protocol ensures a fair comparison across models with different depth parameterizations and output representations.
This setup allows us to isolate the intrinsic capabilities of StereoVGGT by eliminating the influence of the downstream disparity decoder and the inpainting model. 

\noindent
\textbf{1) Can StereoVGGT extract robust stereo features without relying on a dedicated downstream task decoder?}

\begin{figure*}[t]
	\centering
	\includegraphics[width=\linewidth]{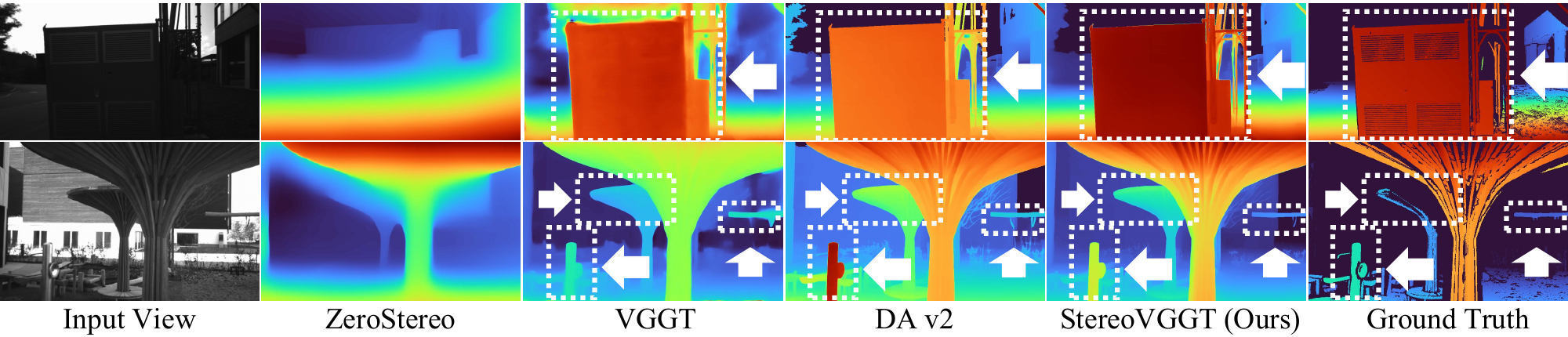}
	\caption{Disparity maps generated by SOTA models on the ETH3D dataset. The models use monocular condition input of binocular cameras, consisting of a left view, focal length, and baseline length. The white boxes highlight the main differences. The disparity maps generated by ZeroStereo exhibit instability and fail to accurately reconstruct some foreground objects. 
	The disparity maps estimated by VGGT exhibit significant limitations in edge-texture fidelity. While DAv2 achieves precise edge localization, it demonstrates inaccuracies in estimating the relative depth ordering and inter-object spatial relationships. In contrast, StereoVGGT succeeds in providing both fine-grained edge estimation and a highly accurate global geometric layout.
	} 
	\label{vis-disp} 
\end{figure*}

\begin{table*}[t]
	\centering
	\setlength{\tabcolsep}{0.5mm}
	\begin{tabular}{c|c|ccccccc|cc|cccc}
		\toprule[1.5 pt]
		\multirow{3}{*}{No.} &
		\multirow{3}{*}{\begin{tabular}[c]{@{}c@{}}DINO \\Weights \end{tabular}}&  & \multicolumn{5}{c}{Camera FOV  $\downarrow$}                           &  &  &      \multirow{3}{*}{SSIM $\uparrow$} &\multicolumn{4}{c}{Disparity Error $\downarrow$}                                               \\ 
		&&  & \multicolumn{2}{c}{FOV x} &  & \multicolumn{2}{c}{FOV y} &  &  & &\multirow{2}{*}{EPE} & \multirow{2}{*}{7 px} & \multirow{2}{*}{9 px} & \multirow{2}{*}{15 px}  \\ 
		&&  & med.        & mean        &  & med.        & mean        &  &  &                      &                       &                       &                 &       \\ \midrule
		1&DINOv2           &  &      80.3       &       84.4      &  &       65.8      &       62.6      &  &   &     \cellcolor{red!20} 0.965  &           20.6           &             69.1          &            62.4           &        39.9        \\ 
		2&DAv2               &  &       73.7      &      77.2       &  &      60.0       &      61.8       &  &   & \cellcolor{red!20}0.965  &         \cellcolor{green!10}7.38            &        \cellcolor{green!10}30.5             &        \cellcolor{green!10}26.4             &       \cellcolor{green!10}11.0         \\
		3&MoGe-2               &  &    \cellcolor{green!10}30.4       &     \cellcolor{green!10}35.1       &  &     \cellcolor{green!10}15.8       &     \cellcolor{green!10}14.2       &  & &   \cellcolor{red!20}0.965    & \cellcolor{orange!20}3.68             &  \cellcolor{orange!20}16.5            &  \cellcolor{orange!20}9.78           &       \cellcolor{orange!20}0.24         \\ 
		4&VGGT              &  &     \cellcolor{orange!20}14.7       & \cellcolor{orange!20}18.6      &  &      \cellcolor{orange!20}7.08      & \cellcolor{orange!20}8.13     &  &  &      \cellcolor{green!10}0.776   &           12.1          &           66.5            &           49.8            &       25.7         \\ \midrule
		5&Ours &  & \cellcolor{red!20}3.88    & \cellcolor{red!20}6.67     &  & \cellcolor{red!20}2.68        & \cellcolor{red!20}5.09       &  &  &        \cellcolor{orange!20}0.964  & \cellcolor{red!20}2.38                & \cellcolor{red!20}10.7               & \cellcolor{red!20}7.27                 & \cellcolor{red!20}0.16      \\
		\bottomrule[1.5 pt]    
	\end{tabular}
	\caption{Impact of weight selection. The weights of the DINO were re-weighted in StereoVGGT. SSIM calculation followed the same configuration as in Figs. 3 and 4. The camera FOV error is calculated following the methods in \citep{moge,vggt}, with specific computational details provided in the Preliminary section.}
	\label{dino-weight}
\end{table*}
\begin{table*}[t]
\centering
\begin{tabular}{cc|cccc|c|cccc}
	\toprule[1.5 pt]
	\multirow{3}{*}{$\lambda_{VGGT}$} & \multirow{3}{*}{$\lambda_{MDE}$} & \multicolumn{4}{c|}{Camera FOV $\downarrow$}                        & \multirow{3}{*}{SSIM $\uparrow$} & \multicolumn{4}{c}{Disparity Error $\downarrow$}                                                                     \\
	&                            & \multicolumn{2}{c}{FOV x} & \multicolumn{2}{c|}{FOV y} &                       & \multirow{2}{*}{EPE} & \multirow{2}{*}{7 px} & \multirow{2}{*}{9 px} & \multirow{2}{*}{15 px} \\
	&                            & med.         & mean       & med.        & mean        &                       &                      &                       &                       &                        \\ \midrule
	0.0                         & 1.0                        & 30.4         & 35.1       & 15.8        & 14.2        & 0.965                 & 3.68                 & 16.5                  & 9.78                  & 0.24                   \\
	0.2         & 0.8           & 29.3 & 35.8 & 17.2 & 20.0 & 0.842 & 12.6 & 75.3 & 56.6 & 25.3  \\
	0.5    & 0.5      &  27.7 & 28.5 & 16.5 & 19.7 & 0.799 & 12.8 & 72.4 & 50.9 & 23.5\\
	0.8        & 0.2      &  26.1 & 21.3 & 13.9 & 13.6 & 0.727 & 12.3 & 79.10 & 48.7 & 24.03 \\
	1.0                         & 0.0                        & 14.7         & 18.6       & 7.08        & 8.13        & 0.776                 & 12.1                 & 66.5                  & 49.8                  & 25.7                   \\ \midrule
	\multicolumn{2}{c|}{Ours}                                 & \cellcolor{red!20}3.88         & \cellcolor{red!20}6.67       & \cellcolor{red!20}2.68        & \cellcolor{red!20}5.09        & \cellcolor{red!20}0.964                 & \cellcolor{red!20}2.38                 & \cellcolor{red!20}10.7                  & \cellcolor{red!20}7.27                  & \cellcolor{red!20}0.16                    \\
	\bottomrule[1.5 pt]    
\end{tabular}
\caption{Ablation study of $\lambda$ selection. For each experiment, a fixed interpolation coefficient is applied uniformly across all network layers.}
\label{simplex-lambda}
\end{table*}
\begin{table*}[t]
	\centering
	\setlength{\tabcolsep}{0.5mm}
	\begin{tabular}{c|ccclc|ccccccc|cc|cccc}
		\toprule[1.5 pt]
		\multirow{3}{*}{No.} &
		\multirow{3}{*}{\begin{tabular}[c]{@{}c@{}}MDE\\ Neck\end{tabular}} & \multicolumn{3}{c}{AA Block}                                     &  &  & \multicolumn{5}{c}{Camera FOV $\downarrow$}                           &  &&\multirow{3}{*}{SSIM $\uparrow$}      & \multicolumn{4}{c}{Disparity Error $\downarrow$}          \\ 
		&& \multirow{2}{*}{GA} & \multirow{2}{*}{FA} &  \multirow{2}{*}{$\alpha$} &  &  & \multicolumn{2}{c}{FOV x} &  & \multicolumn{2}{c}{FOV y} &  &  & &\multirow{2}{*}{EPE} & \multirow{2}{*}{7 px} & \multirow{2}{*}{9 px} & \multirow{2}{*}{15 px} \\ 
		&&                                   &                                 &  &  &  & med.     & mean        &  & med.        & mean        &  &  &                      &                       &                       &     &\\ \midrule
		1& &                \checkmark                   &                   \checkmark               &    -1.0               &  &  &       5.78      &  \cellcolor{green!10}6.70       &  &      4.27       &   \cellcolor{orange!20}4.98      &  &  & \cellcolor{green!10}0.894    &            2.76          &   \cellcolor{orange!20}11.1             &     \cellcolor{orange!20}5.24             &            0.33       \\
		2&\checkmark &                                   &                                  &        0.0           &  &  &  \cellcolor{green!10}4.05       &       6.76      &  &  \cellcolor{green!10}2.73      &      5.19       &  &  & \cellcolor{red!20}0.964   &  \cellcolor{green!10}2.43                &            \cellcolor{red!20}10.7          &            7.80           &            \cellcolor{red!20}0.15          \\
		3&\checkmark &                 \checkmark                  &                \checkmark                  &         1.0          &  &  &      4.59      &      \cellcolor{red!20}6.40       &  &     3.28        &      \cellcolor{red!20}4.79       &  & &  \cellcolor{orange!20}0.939     &           2.65           &   \cellcolor{green!10}11.5           &            \cellcolor{red!20}4.78          &  \cellcolor{orange!20}0.22             \\ 
		4&\checkmark &            \checkmark                       &                        &         0.2          &  &  &  \cellcolor{orange!20}3.93       & \cellcolor{orange!20}6.67      &  &         \cellcolor{orange!20}2.72     &  \cellcolor{green!10}5.09       &&    & \cellcolor{red!20}0.964     &  \cellcolor{orange!20}2.39 &           \cellcolor{red!20}10.7           &           \cellcolor{green!10}7.26            &    \cellcolor{green!10}0.16                \\ 
		5&\checkmark &                                   &             \checkmark                     &          0.2        &  &  & \cellcolor{red!20}3.88        & \cellcolor{orange!20}6.67        &  & \cellcolor{red!20}2.68        & \cellcolor{green!10}5.09        &  &    & \cellcolor{red!20}0.964   & \cellcolor{red!20}2.38                 & \cellcolor{red!20}10.7                & 7.27                  & \cellcolor{green!10}0.16          \\
		\bottomrule[1.5 pt]    
	\end{tabular}
	\caption{Ablation Study on the Neck Component. ``AA" is Alternative Attention (AA) Block in VGGT, ``GA" is Global Attention in VGGT, ``FA" is Frame Attention in VGGT. Given that the integration of the AA blocks is implemented via element-wise subtraction within our pipeline, the hyperparameter $\alpha$ is accordingly assigned a negative value in Experiment No.1. }
	\label{neck-abl}
\end{table*}
\begin{table}[t]
	\centering
	\setlength{\tabcolsep}{0.5mm}
	\begin{tabular}{cccccccccccccc}
		\toprule[1.5 pt]
		\multirow{3}{*}{$\alpha$} &  &  & \multicolumn{5}{c}{Camera FOV $\downarrow$}                           &  &  & \multicolumn{4}{c}{Disparity Error $\downarrow$}                                                      \\ 
		&  &  & \multicolumn{2}{c}{FOV x} &  & \multicolumn{2}{c}{FOV y} &  &  & \multirow{2}{*}{EPE} & \multirow{2}{*}{7 px} & \multirow{2}{*}{9 px} & \multirow{2}{*}{15 px} \\ 
		&  &  & med.        & mean        &  & med.        & mean        &  &  &                      &                       &                       &                        \\ \midrule
		1.0               &  &  &      4.59       &      \cellcolor{red!20}6.40     &  &       3.28      &  \cellcolor{red!20}4.79      &  &  &          2.65            &             11.5          &   \cellcolor{red!20}4.78         &           0.22             \\
		0.8               &  &  &       4.73      &     \cellcolor{orange!20}6.42       &  &       3.46      & \cellcolor{orange!20}4.82      &  &  &          2.58            &          11.74             &  \cellcolor{orange!20}5.47           &   \cellcolor{green!10}0.20             \\
		0.5               &  &  &      3.97       &      \cellcolor{green!10}6.52     &  &       2.91      &      \cellcolor{green!10}4.94       &  &  &          2.43            &            11.26          &             7.36          &   \cellcolor{orange!20}0.17              \\
		0.4               &  &  & \cellcolor{red!20} 3.81       & 6.57        &  & \cellcolor{red!20}2.62        & 4.99        &  &  & \cellcolor{orange!20}2.40                 & 11.07                 & 7.45                  & \cellcolor{red!20}0.16                  \\
		0.3               &  &  & \cellcolor{orange!20}3.84        & 6.62        &  & \cellcolor{orange!20}2.65        & 5.04        &  &  & \cellcolor{red!20}2.38                & \cellcolor{green!10}10.87                 & \cellcolor{green!10}7.25                & \cellcolor{red!20}0.16          \\
		0.2               &  &  & \cellcolor{green!10}3.88        & 6.67        &  & \cellcolor{green!10}2.68        & 5.09        &  &  & \cellcolor{red!20}2.38             & \cellcolor{red!20}10.70       & 7.27                  & \cellcolor{red!20}0.16          \\
		0.1               &  &  &      3.92       &      6.72       &  &      2.70       &      5.14       &  &  &   \cellcolor{green!10}2.41            &   \cellcolor{orange!20}10.74            &            7.29           &            \cellcolor{red!20}0.16     \\
		\bottomrule[1.5 pt]    
	\end{tabular}
	\caption{An ablation study on hyperparameter $\alpha$ performed using the ETH3D dataset. A larger value of $\alpha$ assigns a greater weight to the VGGT neck branch.}
	\label{alpha}
\end{table}
We evaluate StereoVGGT against three distinct categories of SOTA methods to ensure a comprehensive benchmarking. The first category (Generation-based method) denotes the generative framework specifically engineered for disparity estimation. The original ZeroStereo \citep{zerostereo} needs two views as input. We evaluate the quality of the disparity maps generated by its Adaptive Disparity Selection module.
The second category (Monocular Depth Estimation methods) includes established MDE models \citep{unidepth,moge,dav2}, which are frequently employed as feature extraction backbones in stereo matching pipelines. The third category (3D Reconstruction methods) encompasses general-purpose 3D reconstruction models \citep{vggt,fastvggt}. 
Experimental results in Tab. \ref{mde} demonstrate that StereoVGGT achieves superior performance compared to these methods. 
As illustrated in Fig. \ref{vis-disp}, the inherent feature-level degradation in VGGT \citep{vggt} results in a significant loss of fine-grained edge details. Conversely, while DAv2 \citep{dav2} maintains higher semantic clarity, its calibration-agnostic nature---characterized by a lack of explicit camera-geometry awareness---leads to incorrect spatial relationships and distorted depth hierarchies between objects. By successfully integrating the strengths of both paradigms, StereoVGGT mitigates these structural ambiguities, preserving sharp object boundaries while ensuring a geometrically consistent global layout.

\noindent
\textbf{2) Does the proposed EMWM strategy yield superior performance compared to directly using existing weights?} 

We evaluate the impact of different weight initializations by substituting the DINO component of StereoVGGT with weights from DINOv2 \citep{dinov2}, DAv2 \citep{dav2}, MoGe-2 \citep{moge}, and the VGGT \citep{vggt}. 
A comparative analysis of rows 1-3 and row 5 in Tab. \ref{dino-weight} demonstrates that, under nearly equivalent levels of feature-level degradation, the incorporation of more precise camera pose features results in a substantial reduction in the EPE. This finding underscores the pivotal role of geometric camera priors in achieving high-precision disparity estimation. Furthermore, results from Experiment No. 4 indicate that, despite the inherent capacity of the VGGT for robust camera pose representation, it yields suboptimal performance in precise disparity estimation. This limitation is primarily attributable to the geometric detail degradation during the feature construction phase, as quantitatively corroborated by the diminished SSIM metrics. As evidenced by the results in Experiment No. 5, our proposed EMWM achieves superior performance. The strategy successfully preserves the structural integrity of the feature maps while simultaneously incorporating camera-aware geometry. This synergy effectively optimizes the representation for stereo matching, yielding superior accuracy and robustness in disparity estimation.

\begin{table}[t]
	\centering
	\begin{minipage}[t]{0.37\linewidth}
		\centering
		\setlength{\tabcolsep}{0.5mm}
		\begin{tabular}{c|cc|c}
			\toprule[1.5 pt]    
			Model & Time (s) $\downarrow$& Param. (M) $\downarrow$& EPE (px) $\downarrow$\\ \midrule
			MoGe-2      &   0.156   &    326.21    &   6.74  \\
			DA v2       &   0.328   &    335.32    &  5.88   \\
			VGGT        &   0.197    &    1256.54    &  13.07   \\ \midrule
			Ours  &  0.122    &   937.14    &  2.71   \\
			\bottomrule[1.5 pt]  
		\end{tabular}
		\captionof{table}{To evaluate computational cost, we performed inference for each model on the KITTI dataset using the same experimental setup as in Sec. \ref{ablations}.}
		\label{cost}
	\end{minipage}\hfill
	\begin{minipage}[t]{0.60\linewidth}
		\centering
		\setlength{\tabcolsep}{1.0mm}
		\begin{tabular}{c|cccc|ccc}
			\toprule[1.5 pt]    
			\multirow{2}{*}{Methods}&  & \multicolumn{2}{c}{FOV x $\downarrow$} &  &  & \multicolumn{2}{c}{FOV y $\downarrow$} \\ 
			&  & med.        & mean        &  &  & med.        & mean        \\ \midrule
			fastVGGT          &  &      \cellcolor{orange!20}12.71      &       \cellcolor{orange!20}16.58     &  &  &      \cellcolor{orange!20}6.83       &     \cellcolor{orange!20}8.13       \\ 
			VGGT              &  &      \cellcolor{green!10}12.85       &      \cellcolor{green!10}16.60       &  &  &       \cellcolor{green!10}7.08      &    \cellcolor{orange!20}8.13     \\
			DAv2 $^\dagger$&  &      73.81    &      73.79       &  &  &       60.14      &        61.88     \\
			MoGe-2 $^\dagger$&  &      32.29       &      39.52       &  &  &       31.79      &        30.02     \\
			VGGT $^\dagger$      &  &      14.97       &      18.08       &  &  &       7.19     &        \cellcolor{green!10}8.42    \\
			\midrule
			Ours &  &     \cellcolor{red!20}3.88        &       \cellcolor{red!20}6.67      &  &  &     \cellcolor{red!20}2.68        &      \cellcolor{red!20}5.09     \\
			\bottomrule[1.5 pt]    
		\end{tabular}
		\captionof{table}{Evaluation of left-view camera FOV in degrees. The dataset evaluated in this study was ETH3D binocular stereo datasets. Each method receives only the left-view image as input and outputs the intrinsic parameters of the left camera. $\dagger$ indicate the methods utilize the identical camera-pose solver as that integrated into StereoVGGT.}
		\label{fov}
	\end{minipage}
\end{table}

\noindent
\textbf{3) Is Entropy-Minimized Weight Merging an effective strategy for Selecting $\lambda$?}

To investigate whether entropy minimization provides an effective criterion for selecting the merging coefficient, we compare the proposed strategy against a set of fixed interpolation points on the simplex $\lambda_{\mathrm{VGGT}}+\lambda_{\mathrm{MDE}}=1$, as summarized in Tab. \ref{simplex-lambda}. Specifically, different combinations of $\lambda_{\mathrm{VGGT}}$ and $\lambda_{\mathrm{MDE}}$ are evaluated while keeping all other settings unchanged. The results demonstrate that the entropy-minimized solution yields the best overall performance, indicating that entropy minimization serves as a reliable proxy for identifying high-quality merged weights.

\noindent
\textbf{4) Does the proposed feature neck outperform using the feature neck from MDE or VGGT directly?}

This ablation study investigates the relative contributions of the MDE neck and the VGGT neck to the overall system performance. The corresponding results are summarized in Tab. \ref{neck-abl}. The condition $\alpha=-1.0$ corresponds to a baseline configuration where the feature neck is exclusively composed of the frozen AA blocks from VGGT, with no feature modulation being performed by the MDE branch. 
In contrast, Experiment No. 2 utilizes the MDE neck in isolation, excluding the VGGT neck from the architecture. Under these specific configurations, both the camera FOV estimation and the resulting disparity exhibit deviations from the ground truth. 
Experiments No. 3-5 explore various integration strategies for the two branches, revealing that the configuration in Experiment No. 5 achieves the superior disparity estimation performance. 
Notably, this result also demonstrates that balancing camera-specific knowledge with feature preservation capacity facilitates the expression of more robust stereo vision priors.

\noindent
\textbf{5) Impact of Hyperparameter $\alpha$.} 

This part presents an ablation study on hyperparameter $\alpha$, with the corresponding results provided in Tab. \ref{alpha}. Experiments indicate that the mean camera FOV is estimated most accurately when $\alpha$ is set to 1.0. Furthermore, the median FOV error exhibits a monotonic decrease when $\alpha \geq 0.4$.
However, an excessively large weight for the VGGT neck can degrade the performance of both camera pose and disparity estimation, likely due to image degradation induced by an overemphasis on this branch. 
In our experimental setup, the EPE is minimized when $\alpha$ is 0.2 or 0.3. Notably, the camera pose error at these values is also competitive with known SOTA levels, providing further evidence for the correlation between camera knowledge and disparity estimation capability. This suggests that an optimal value of $\alpha$ within the range of 0.2 to 0.4 may exist, striking a balance between camera knowledge and feature representation to achieve even higher disparity estimation accuracy. Future work could explore learning-based methods to adaptively determine this hyperparameter.

\subsection{Computational Cost}
This section compares the computational cost across the different models. All experiments were conducted on a single NVIDIA RTX 3090 GPU. The model inputs consisted of the left-view images from the KITTI dataset, along with the focal length and baseline length from the KITTI camera parameters. The output was a disparity map, and the results are presented in Tab. \ref{cost}. The time reported in the table represents the average duration over 50 inference runs. 
Despite incorporating a substantial number of parameters, StereoVGGT does not suffer from a significant increase in inference time, as a large portion of the parameters from the MDE model and VGGT are loaded but remain inactive during inference. Remarkably, StereoVGGT achieves a faster inference speed than DAv2, MoGe-2, and VGGT. 

\subsection{Disparity Perception under Ill-posed Conditions}
Ill-posed conditions refers to challenging scenarios such as thin structures, reflective surfaces, and textureless or weakly-textured regions, where achieving accurate disparity estimation is particularly difficult. The ETH3D dataset \citep{eth3d} is specifically designed to include these ill-posed conditions. Building upon the results shown in Tab. \ref{mde}, this section offers a deeper analysis of the model's performance on this dataset.

\begin{figure*}[h]
	\centering
	\includegraphics[width=\linewidth]{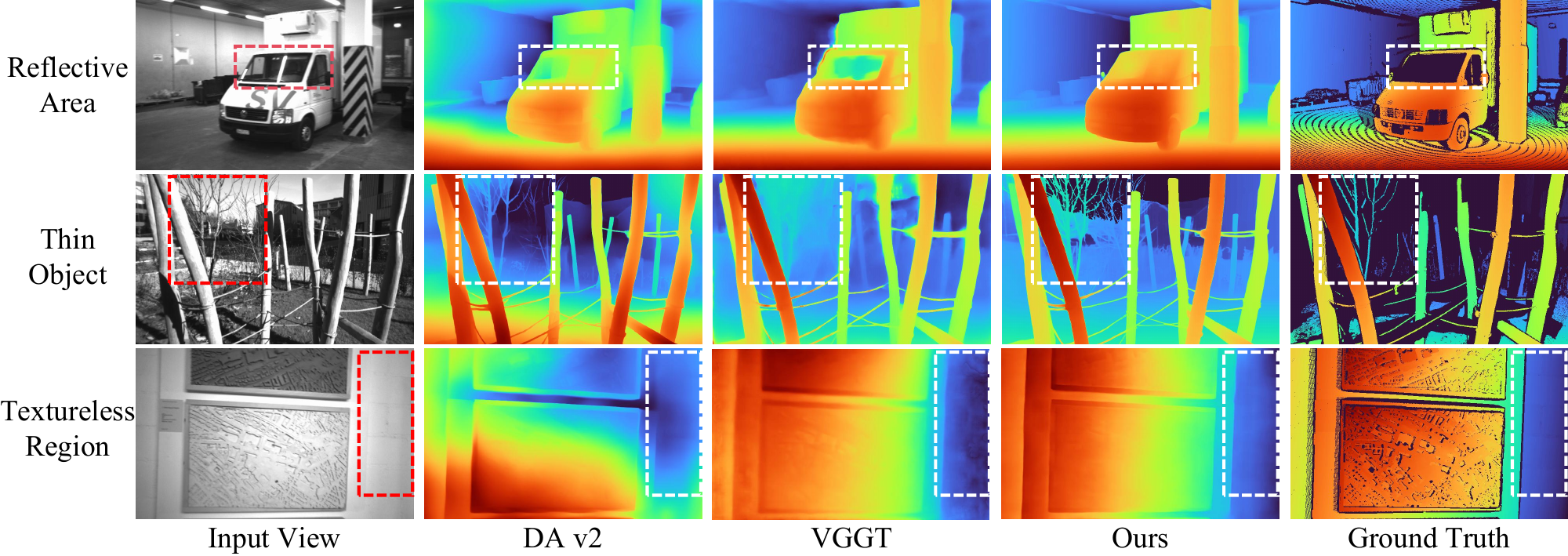} 
	\caption{Disparity map comparisons across state-of-the-art models on challenging ill-posed regions. The models use monocular condition input of binocular cameras, consisting of a left view, focal length, and baseline length. The red boxes and white boxes highlight the main differences. 
	} 
	\label{ill} 
\end{figure*}

In addition to the quantitative analysis provided in the main manuscript, we present a visual comparison of different types of ill-posed regions in Fig. \ref{ill}. 
Specifically, MDE models, exemplified by DAv2 \citep{dav2}, frequently confuse the spatial relationships between scene objects and the imaging plane, whereas VGGT \citep{vggt} often produces blurred object boundaries. In contrast, StereoVGGT achieves both fine-grained edge estimation and accurate relative distance estimation for objects.

\section{Conclusion}
\label{sec:conclusion}

We introduce StereoVGGT, a training-free VGGT designed to serve as a backbone for stereo vision tasks. This research is motivated by the absence of explicit camera knowledge in current stereo vision backbones. The omission of such knowledge imposes a performance bottleneck when these backbones are deployed in downstream applications. To bridge this gap, we developed a novel architecture derived from the VGGT. However, the vanilla VGGT suffers from feature-level structural degradation, rendering it suboptimal for high-precision stereo feature extraction. StereoVGGT effectively integrates geometric camera knowledge while preserving fine-grained feature integrity. Our results provide the first empirical validation that a 3D foundation model can serve as an effective backbone for stereo vision. 

\section{Limitations and Future Work}

With the rapid advancement of 3D foundation models, future work could explore leveraging stronger 3D foundation models to further enhance stereo vision representations. However, 
StereoVGGT introduces additional computational overhead due to the integration of both a 3D foundation model and an MDE model, resulting in a larger parameter footprint compared with conventional PVM- or MDE-based stereo backbones. 
Future extensions that incorporate stronger foundation models may inevitably introduce increased memory consumption and deployment overhead. 
Therefore, developing parameter-efficient fusion strategies and lightweight adaptation mechanisms remains an important direction for improving the scalability.

\bibliography{aaai2027-appendix}

@String(PAMI = {IEEE Trans. Pattern Anal. Mach. Intell.})

@String(IJCV = {Int. J. Comput. Vis.})

@String(CVPR = {Proc. IEEE/CVF Conf. Comput. Vis. Pattern Recog.})

@String(ICCV = {Proc. IEEE/CVF Int. Conf. Comput. Vis.})

@String(ECCV = {Proc. Eur. Conf. Comput. Vis.})

@String(NIPS = {Adv. Neural Inform. Process. Syst.})

@String(TIP = {IEEE Trans. Image Process.})

@String(ICLR = {Int. Conf. Learn. Represent.})

@String(AAAI = {Proc. AAAI Conf. Artif. Intell.})

@String(CVPRW = {Proc. IEEE/CVF Conf. Comput. Vis. Pattern Recog. Worksh.})

@String(TDV = {Int. Conf. 3D Vis.})

@InProceedings{obukhov2025fourth,
	author    = {Obukhov, Anton and Poggi, Matteo and Tosi, Fabio and Arora, Ripudaman Singh and Spencer, Jaime and Russell, Chris and Hadfield, Simon and Bowden, Richard and Wang, Shuaihang and Ma, Zhenxin and Chen, Weijie and Xu, Baobei and Sun, Fengyu and Xie, Di and Zhu, Jiang and Lavreniuk, Mykola and Guan, Haining and Wu, Qun and Zeng, Yupei and Lu, Chao and Wang, Huanran and Zhou, Guangyuan and Zhang, Haotian and Wang, Jianxiong and Rao, Qiang and Wang, Chunjie and Liu, Xiao and Lou, Zhiqiang and Jiang, Hualie and Chen, Yihao and Xu, Rui and Tan, Minglang and Qin, Zihan and Mao, Yifan and Liu, Jiayang and Xu, Jialei and Yang, Yifan and Zhao, Wenbo and Jiang, Junjun and Liu, Xianming and Zhao, Mingshuai and Ming, Anlong and Chen, Wu and Xue, Feng and Yu, Mengying and Gao, Shida and Wang, Xiangfeng and Omotara, Gbenga and Farag, Ramy and Demby's, Jacket and Tousi, Seyed Mohamad Ali and DeSouza, Guilherme N. and Yang, Tu{\^{}\'a}n Anh and Nguy\~{\^{e}}n, Minh Quang and Tr\^{a}n, Thi\^{e}n Ph\'{u}c and Luginov, Albert and Shahzad, Muhammad},
	title     = {The fourth monocular depth estimation challenge},
	booktitle = CVPRW,
	month     = {June},
	year      = {2025},
	pages     = {6228--6241},
	doi       = {10.1109/CVPRW67362.2025.00615}
}

@article{pvm-notion1,
	author  = {Xin, Yi and Yang, Jianjiang and Luo, Siqi and Du, Yuntao and Qin, Qi and Chen, Haoxing and Cen, Kangrui and He, Yangfan and Fu, Bin and Cao, Yuewen and He, Junjun and Yang, Xiaokang and Zhai, Guangtao and Yang, Ming-Hsuan and Liu, Xiaohong},
	title   = {Parameter-efficient fine-tuning for pre-trained vision models: A survey and benchmark},
	journal = IJCV,
	volume  = {134},
	number  = {6},
	pages   = {304},
	year    = {2026},
	doi     = {10.1007/s11263-026-02864-6},
	note    = {Published online: 03 June 2026}
}

@inproceedings{dust3r,
	title={{DUSt3R}: Geometric 3D vision made easy},
	author={Wang, Shuzhe and Leroy, Vincent and Cabon, Yohann and Chidlovskii, Boris and Revaud, Jerome},
	booktitle=CVPR,
	pages={20697--20709},
	year={2024}
}

@inproceedings{croco,
	title={{CroCo} v2: Improved cross-view completion pre-training for stereo matching and optical flow},
	author={Weinzaepfel, Philippe and Lucas, Thomas and Leroy, Vincent and Cabon, Yohann and Arora, Vaibhav and Br{\'e}gier, Romain and Csurka, Gabriela and Antsfeld, Leonid and Chidlovskii, Boris and Revaud, J{\'e}r{\^o}me},
	booktitle=ICCV,
	pages={17969--17980},
	year={2023}
}

@inproceedings{vggt,
	title={{VGGT}: Visual geometry grounded transformer},
	author={Wang, Jianyuan and Chen, Minghao and Karaev, Nikita and Vedaldi, Andrea and Rupprecht, Christian and Novotny, David},
	booktitle=CVPR,
	pages={5294--5306},
	year={2025}
}

@article{promptstereo,
	title={{PromptStereo}: Zero-shot stereo matching via structure and motion prompts},
	author={Wang, Xianqi and Yang, Hao and Wang, Hangtian and Cheng, Junda and Xu, Gangwei and Lin, Min and Yang, Xin},
	journal=CVPR,
	year={2026}
}

@inproceedings{dpt,
	title={Vision transformers for dense prediction},
	author={Ranftl, Ren{\'e} and Bochkovskiy, Alexey and Koltun, Vladlen},
	booktitle=ICCV,
	pages={12179--12188},
	year={2021}
}

@article{pei2019effects,
	title={Effects of image degradation and degradation removal to CNN-based image classification},
	author={Pei, Yanting and Huang, Yaping and Zou, Qi and Zhang, Xingyuan and Wang, Song},
	journal=PAMI,
	volume={43},
	number={4},
	pages={1239--1253},
	year={2019},
	publisher={IEEE}
}

@inproceedings{wang2020deep,
	title={Deep degradation prior for low-quality image classification},
	author={Wang, Yang and Cao, Yang and Zha, Zheng-Jun and Zhang, Jing and Xiong, Zhiwei},
	booktitle=CVPR,
	pages={11049--11058},
	year={2020}
}

@article{wang2004image,
	title={Image quality assessment: from error visibility to structural similarity},
	author={Wang, Zhou and Bovik, Alan C and Sheikh, Hamid R and Simoncelli, Eero P},
	journal=TIP,
	volume={13},
	number={4},
	pages={600--612},
	year={2004},
	publisher={IEEE}
}

@inproceedings{zeiler2014visualizing,
	title={Visualizing and understanding convolutional networks},
	author={Zeiler, Matthew D and Fergus, Rob},
	booktitle=ECCV,
	pages={818--833},
	year={2014},
	organization={Springer}
}

@inproceedings{taskvector,
	title={Editing models with task arithmetic},
	author={Ilharco, Gabriel and Ribeiro, Marco Tulio and Wortsman, Mitchell and Schmidt, Ludwig and Hajishirzi, Hannaneh and Farhadi, Ali},
	booktitle=ICLR,
	year=2023
}

@inproceedings{xu2024training,
	title={Training-free pretrained model merging},
	author={Xu, Zhengqi and Yuan, Ke and Wang, Huiqiong and Wang, Yong and Song, Mingli and Song, Jie},
	booktitle=CVPR,
	pages={5915--5925},
	year={2024}
}

@inproceedings{bau2017network,
	title={Network dissection: Quantifying interpretability of deep visual representations},
	author={Bau, David and Zhou, Bolei and Khosla, Aditya and Oliva, Aude and Torralba, Antonio},
	booktitle=CVPR,
	pages={6541--6549},
	year={2017}
}

@misc{fastvggt,
	title={{FastVGGT}: Training-free acceleration of visual geometry transformer},
	author={Shen, You and Zhang, Zhipeng and Qu, Yansong and Cao, Liujuan},
	eprint={2509.02560},
	archivePrefix={arXiv},
	year={2025}
}

@misc{mochav2,
	title={Motif channel opened in a white-box: Stereo matching via motif correlation graph},
	author={Chen, Ziyang and Zhang, Yongjun and Li, Wenting and Wang, Bingshu and Zhao, Yong and Chen, CL},
	eprint={2411.12426},
	archivePrefix={arXiv},
	year={2024}
}

@article{dav2,
	title={Depth Anything v2},
	author={Yang, Lihe and Kang, Bingyi and Huang, Zilong and Zhao, Zhen and Xu, Xiaogang and Feng, Jiashi and Zhao, Hengshuang},
	journal=NIPS,
	volume={37},
	pages={21875--21911},
	year={2024}
}

@inproceedings{monster,
	title={{MonSter}: Marry monodepth to stereo unleashes power},
	author={Cheng, Junda and Liu, Longliang and Xu, Gangwei and Wang, Xianqi and Zhang, Zhaoxing and Deng, Yong and Zang, Jinliang and Chen, Yurui and Cai, Zhipeng and Yang, Xin},
	booktitle=CVPR,
	pages={6273--6282},
	year={2025}
}

@inproceedings{bridgedepth,
	title={{BridgeDepth}: Bridging monocular and stereo reasoning with latent alignment},
	author={Guan, Tongfan and Guo, Jiaxin and Wang, Chen and Liu, Yun-Hui},
	booktitle=CVPR,
	pages={27681--27691},
	year={2025}
}

@article{midas,
	title={Towards robust monocular depth estimation: Mixing datasets for zero-shot cross-dataset transfer},
	author={Ranftl, Ren{\'e} and Lasinger, Katrin and Hafner, David and Schindler, Konrad and Koltun, Vladlen},
	journal=PAMI,
	volume={44},
	number={3},
	pages={1623--1637},
	year={2020},
	publisher={IEEE}
}

@inproceedings{selective,
	title={Selective-stereo: Adaptive frequency information selection for stereo matching},
	author={Wang, Xianqi and Xu, Gangwei and Jia, Hao and Yang, Xin},
	booktitle=CVPR,
	pages={19701--19710},
	year={2024}
}

@misc{moge,
	title={{MoGe-2}: Accurate monocular geometry with metric scale and sharp details},
	author={Wang, Ruicheng and Xu, Sicheng and Dong, Yue and Deng, Yu and Xiang, Jianfeng and Lv, Zelong and Sun, Guangzhong and Tong, Xin and Yang, Jiaolong},
	eprint={2507.02546},
	archivePrefix={arXiv},
	year={2025}
}

@inproceedings{marigold,
	title={Repurposing diffusion-based image generators for monocular depth estimation},
	author={Ke, Bingxin and Obukhov, Anton and Huang, Shengyu and Metzger, Nando and Daudt, Rodrigo Caye and Schindler, Konrad},
	booktitle=CVPR,
	pages={9492--9502},
	year={2024}
}

@misc{dinov2,
	title={{DINOv2}: Learning robust visual features without supervision},
	author={Oquab, Maxime and Darcet, Timoth{\'e}e and Moutakanni, Th{\'e}o and Vo, Huy and Szafraniec, Marc and Khalidov, Vasil and Fernandez, Pierre and Haziza, Daniel and Massa, Francisco and El-Nouby, Alaaeldin and others},
	eprint={2304.07193},
	archivePrefix={arXiv},
	year={2023}
}

@article{vitas,
	title={Playing to vision foundation model's strengths in stereo matching},
	author={Liu, Chuang-Wei and Chen, Qijun and Fan, Rui},
	journal={IEEE Transactions on Intelligent Vehicles},
	year={2024},
	publisher={IEEE}
}

@article{fdn,
	title={Feature distribution normalization network for multi-view stereo},
	author={Chen, Ziyang and Zhao, Yang and He, Junling and Lu, Yujie and Cui, Zhongwei and Li, Wenting and Zhang, Yongjun},
	journal={The Visual Computer},
	volume={41},
	number={1},
	pages={409--421},
	year={2025},
	publisher={Springer}
}

@inproceedings{los,
	title={{LoS}: Local structure-guided stereo matching},
	author={Li, Kunhong and Wang, Longguang and Zhang, Ye and Xue, Kaiwen and Zhou, Shunbo and Guo, Yulan},
	booktitle=CVPR,
	pages={19746--19756},
	year={2024}
}

@inproceedings{mobilenetv2,
	title={{MobileNetV2}: Inverted residuals and linear bottlenecks},
	author={Sandler, Mark and Howard, Andrew and Zhu, Menglong and Zhmoginov, Andrey and Chen, Liang-Chieh},
	booktitle=CVPR,
	pages={4510--4520},
	year={2018}
}

@article{levenberg,
	title={The Levenberg-Marquardt algorithm},
	author={Ranganathan, Ananth},
	journal={Tutoral on LM algorithm},
	volume={11},
	number={1},
	pages={101--110},
	year={2004}
}

@inproceedings{mocha,
	title={{MoCha}-stereo: Motif channel attention network for stereo matching},
	author={Chen, Ziyang and Long, Wei and Yao, He and Zhang, Yongjun and Wang, Bingshu and Qin, Yongbin and Wu, Jia},
	booktitle=CVPR,
	pages={27768--27777},
	year={2024}
}

@inproceedings{mono2stereo,
	title={{Mono2Stereo}: A benchmark and empirical study for stereo conversion},
	author={Yu, Songsong and Chen, Yuxin and Qi, Zhongang and Xie, Zeke and Wang, Yifan and Wang, Lijun and Shan, Ying and Lu, Huchuan},
	booktitle=CVPR,
	pages={21847--21856},
	year={2025}
}

@misc{stereocrafter,
	title={{StereoCrafter}: Diffusion-based generation of long and high-fidelity stereoscopic 3D from monocular videos},
	author={Zhao, Sijie and Hu, Wenbo and Cun, Xiaodong and Zhang, Yong and Li, Xiaoyu and Kong, Zhe and Gao, Xiangjun and Niu, Muyao and Shan, Ying},
	eprint={2409.07447},
	archivePrefix={arXiv},
	year={2024}
}

@inproceedings{depthcrafter,
	author      = {Hu, Wenbo and Gao, Xiangjun and Li, Xiaoyu and Zhao, Sijie and Cun, Xiaodong and Zhang, Yong and Quan, Long and Shan, Ying},
	title       = {{DepthCrafter}: Generating consistent long depth sequences for open-world videos},
	booktitle   = CVPR,
	year        = {2025}
}

@article{arbelaez2010contour,
	title={Contour detection and hierarchical image segmentation},
	author={Arbelaez, Pablo and Maire, Michael and Fowlkes, Charless and Malik, Jitendra},
	journal=PAMI,
	volume={33},
	number={5},
	pages={898--916},
	year={2010},
	publisher={IEEE}
}

@article{rethinking,
	title={Rethinking low-light knowledge for pedestrian detection in nighttime conditions},
	author={Chen, Ziyang and Yao, He and Li, Wenting and Long, Wei and Zhang, Yongjun},
	journal={Engineering Applications of Artificial Intelligence},
	volume={160},
	pages={111979},
	year={2025},
	publisher={Elsevier}
}

@article{sat,
	title={Surface depth estimation from multi-view stereo satellite images with distribution contrast network},
	author={Chen, Ziyang and Li, Wenting and Cui, Zhongwei and Zhang, Yongjun},
	journal={IEEE Journal of Selected Topics in Applied Earth Observations and Remote Sensing},
	year={2024},
	publisher={IEEE}
}

@inproceedings{raft,
	title={{RAFT}-stereo: Multilevel recurrent field transforms for stereo matching},
	author={Lipson, Lahav and Teed, Zachary and Deng, Jia},
	booktitle=TDV,
	pages={218--227},
	year={2021},
	organization={IEEE}
}

@misc{mono-stereo,
	title={{Mono2Stereo}: Monocular knowledge transfer for enhanced stereo matching},
	author={Wang, Yuran and Liang, Yingping and Li, Hesong and Fu, Ying},
	eprint={2411.09151},
	archivePrefix={arXiv},
	year={2024}
}

@inproceedings{aio,
	title={All-in-One: Transferring vision foundation models into stereo matching},
	author={Zhou, Jingyi and Zhang, Haoyu and Yuan, Jiakang and Ye, Peng and Chen, Tao and Jiang, Hao and Chen, Meiya and Zhang, Yangyang},
	booktitle=AAAI,
	volume={39},
	number={10},
	pages={10797--10805},
	year={2025}
}

@inproceedings{unidepth,
	title={{UniDepth}: Universal monocular metric depth estimation},
	author={Piccinelli, Luigi and Yang, Yung-Hsu and Sakaridis, Christos and Segu, Mattia and Li, Siyuan and Van Gool, Luc and Yu, Fisher},
	booktitle=CVPR,
	pages={10106--10116},
	year={2024}
}

@inproceedings{igev,
	title={Iterative geometry encoding volume for stereo matching},
	author={Xu, Gangwei and Wang, Xianqi and Ding, Xiaohuan and Yang, Xin},
	booktitle=CVPR,
	pages={21919--21928},
	year={2023}
}

@article{igev++,
	title={{IGEV}++: Iterative multi-range geometry encoding volumes for stereo matching},
	author={Xu, Gangwei and Wang, Xianqi and Zhang, Zhaoxing and Cheng, Junda and Liao, Chunyuan and Yang, Xin},
	journal=PAMI,
	year={2025},
	publisher={IEEE}
}

@article{gip,
	title={{GIP}-Stereo: Geometry-aware information propagation network for stereo matching},
	author={Zhao, Yang and Chen, Ziyang and He, Junling and Li, Wenting and Xiao, Yao and Tian, Chunwei and Zhang, Yongjun},
	journal={Knowledge-Based Systems},
	pages={114062},
	year={2025},
	publisher={Elsevier}
}

@article{fov,
	title={Straight lines have to be straight},
	author={Devernay, Frederic and Faugeras, Olivier},
	journal={Machine Vision and Applications},
	volume={13},
	number={1},
	pages={14--24},
	year={2001},
	publisher={Springer}
}

@inproceedings{kitti,
	author = {Moritz Menze and Christian Heipke and Andreas Geiger},
	title = {Joint 3D estimation of vehicles and scene flow},
	booktitle = {ISPRS Workshop on Image Sequence Analysis (ISA)},
	year = {2015}
}

@inproceedings{sceneflow,
	title={A large dataset to train convolutional networks for disparity, optical flow, and scene flow estimation},
	author={Mayer, Nikolaus and Ilg, Eddy and Hausser, Philip and Fischer, Philipp and Cremers, Daniel and Dosovitskiy, Alexey and Brox, Thomas},
	booktitle=CVPR,
	pages={4040--4048},
	year={2016}
}

@inproceedings{lee2021ctrlc,
	title={CTRL-C: Camera Calibration TRansformer With Line-Classification},
	author={Lee, Jinwoo and Go, Hyunsung and Lee, Hyunjoon and Cho, Sunghyun and Sung, Minhyuk and Kim, Junho},
	booktitle=ICCV,
	pages={16228--16237},
	year={2021}
}

@inproceedings{patel2025camerahmr,
	title={{CameraHMR}: Aligning People with Perspective},
	author={Patel, Priyanka and Black, Michael J.},
	booktitle=TDV,
	pages={1562--1571},
	year={2025}
}

@article{long2026,
	title={Leveraging negative correlation for full-range self-attention in vision transformers},
	author={Long, Wei and Chen, Ziyang and Li, Wenting and Zhang, Yongjun and Yao, He and Peng, Jiaxin and Cui, Zhongwei},
	journal={Pattern Recognition},
	volume={169},
	pages={111899},
	year={2026},
	publisher={Elsevier}
}

@misc{mvggt,
	title={{MVGGT}: Multimodal visual geometry grounded transformer for multiview 3D referring expression segmentation},
	author={Wu, Changli and Wang, Haodong and Ji, Jiayi and Yao, Yutian and Du, Chunsai and Kang, Jihua and Fu, Yanwei and Cao, Liujuan},
	eprint={2601.06874},
	archivePrefix={arXiv},
	year={2026}
}

@inproceedings{omega,
	title={{VGGT}-Omega},
	author={Wang, Jianyuan and Chen, Minghao and Zhang, Shangzhan and Karaev, Nikita and Sch{\"o}nberger, Johannes and Labatut, Patrick and Bojanowski, Piotr and Novotny, David and Vedaldi, Andrea and Rupprecht, Christian},
	booktitle=CVPR,
	year={2026}
}

@inproceedings{long2025progressive,
	title={Progressive focused transformer for single image super-resolution},
	author={Long, Wei and Zhou, Xingyu and Zhang, Leheng and Gu, Shuhang},
	booktitle=CVPR,
	pages={2279--2288},
	year={2025}
}

@inproceedings{eth3d,
	title={A multi-view stereo benchmark with high-resolution images and multi-camera videos},
	author={Schops, Thomas and Schonberger, Johannes L and Galliani, Silvano and Sattler, Torsten and Schindler, Konrad and Pollefeys, Marc and Geiger, Andreas},
	booktitle=CVPR,
	pages={3260--3269},
	year={2017}
}

@inproceedings{stereodiffusion,
	title={{StereoDiffusion}: Training-free stereo image generation using latent diffusion models},
	author={Wang, Lezhong and Frisvad, Jeppe Revall and Jensen, Mark Bo and Bigdeli, Siavash Arjomand},
	booktitle=CVPR,
	pages={7416--7425},
	year={2024}
}

@inproceedings{zerostereo,
	title={{ZeroStereo}: Zero-shot stereo matching from single images},
	author={Wang, Xianqi and Yang, Hao and Xu, Gangwei and Cheng, Junda and Lin, Min and Deng, Yong and Zang, Jinliang and Chen, Yurui and Yang, Xin},
	booktitle=ICCV,
	year={2025}
}

@inproceedings{imagenet,
	title={ImageNet: A large-scale hierarchical image database},
	author={Deng, Jia and Dong, Wei and Socher, Richard and Li, Li-Jia and Li, Kai and Fei-Fei, Li},
	booktitle=CVPR,
	pages={248--255},
	year={2009},
	organization={IEEE}
}

@article{sgm,
	title={Stereo processing by semiglobal matching and mutual information},
	author={Hirschmuller, Heiko},
	journal=PAMI,
	volume={30},
	number={2},
	pages={328--341},
	year={2007},
	publisher={IEEE}
}

@inproceedings{middlebury,
	title={High-resolution stereo datasets with subpixel-accurate ground truth},
	author={Scharstein, Daniel and Hirschm{\"u}ller, Heiko and Kitajima, York and Krathwohl, Greg and Ne{\v{s}}i{\'c}, Nera and Wang, Xi and Westling, Porter},
	booktitle={German conference on pattern recognition},
	pages={31--42},
	year={2014},
	organization={Springer}
}

@inproceedings{idesplat,
	title={{IDESplat}: Iterative depth probability estimation for generalizable 3D Gaussian splatting},
	author={Long, Wei and Wu, Haifeng and Jiang, Shiyin and Zhang, Jinhua and Ji, Xinchun and Gu, Shuhang},
	booktitle=CVPR,
	year={2026}
}

@inproceedings{zhang2018unreasonable,
	title={The unreasonable effectiveness of deep features as a perceptual metric},
	author={Zhang, Richard and Isola, Phillip and Efros, Alexei A and Shechtman, Eli and Wang, Oliver},
	booktitle=CVPR,
	pages={586--595},
	year={2018}
}

@inproceedings{dlnr,
	title={High-frequency stereo matching network},
	author={Zhao, Haoliang and Zhou, Huizhou and Zhang, Yongjun and Chen, Jie and Yang, Yitong and Zhao, Yong},
	booktitle=CVPR,
	pages={1327--1336},
	year={2023}
}

@inproceedings{xie2015holistically,
	title={Holistically-nested edge detection},
	author={Xie, Saining and Tu, Zhuowen},
	booktitle=ICCV,
	pages={1395--1403},
	year={2015}
}

@inproceedings{liu2017richer,
	title={Richer convolutional features for edge detection},
	author={Liu, Yun and Cheng, Ming-Ming and Hu, Xiaowei and Wang, Kai and Bai, Xiang},
	booktitle=CVPR,
	pages={3000--3009},
	year={2017}
}
\bibliographystyle{bibstyle}
	
\end{document}